\documentclass[11pt,a4paper]{article}
\usepackage[hyperref]{acl2020}
\usepackage{times}
\usepackage{latexsym}
\usepackage{booktabs}
\usepackage{amssymb}
\usepackage{amsmath}
\usepackage{comment}
\usepackage{url}
\usepackage{enumitem}
\usepackage{tikz}
\usepackage{array}
\usepackage{xspace}
\usepackage{etoolbox}
\usepackage{patchcmd}
\def\checkmark{\tikz\fill[scale=0.3](0,.35) -- (.25,0) -- (1,.7) -- (.25,.15) -- cycle;} 

\usepackage[nameinlink]{cleveref}
\crefformat{section}{\S#2#1#3} % see manual of cleveref, section 8.2.1
\crefformat{subsection}{\S#2#1#3}
\crefformat{subsubsection}{\S#2#1#3}

\newcommand{\red}[1]{\textcolor{red}{#1}}
\newcommand{\blue}[1]{\textcolor{blue}{#1}}

\newcommand{\ie}{{\em i.e.,}\xspace}
\newcommand{\eg}{{\em e.g.,}\xspace}

\newcommand{\Ni}{({\em i})~}
\newcommand{\Nii}{({\em ii})~}
\newcommand{\Niii}{({\em iii})~}

\newcommand{\ps}{\rho}

\newcommand{\concat}{{\sc Concat}}
\newcommand{\sts}{{\sc Sen2Sen}}
\newcommand{\anap}{{\sc Anaph}}
\newcommand{\han}{{\sc Han}}
\newcommand{\san}{{\sc San}}

\DeclareMathAlphabet{\pazocal}{OMS}{zplm}{m}{n}
\DeclareMathAlphabet{\pazocal}{OMS}{zplm}{m}{n}

\aclfinalcopy % Uncomment this line for the final submission
\title{Can Your Context-Aware MT System Pass the DiP Benchmark Tests? : \\ Evaluation Benchmarks for \textbf{Di}scourse \textbf{P}henomena in  Machine Translation}

\author{Prathyusha Jwalapuram$^*$, Barbara Rychalska$^\dagger$, Shafiq Joty$^*$$^\S$, and Dominika Basaj$^\dagger$\\
  $^*$Nanyang Technological University, Singapore \\
  $^\S$Salesforce Research Asia, Singapore \\
  $^\dagger$Warsaw University of Technology, Faculty of Mathematics and Information Science\\
  $^*${\tt\{jwal0001,srjoty\}@ntu.edu.sg} \\
  $^\dagger${\tt\{b.rychalska,dbasaj\}@mini.pw.edu.pl} \\
\\}

\date{}

\begin{document}
\maketitle
\begin{abstract}
  Despite increasing instances of machine translation {(MT)} systems including contextual information, the evidence for translation quality improvement is sparse, especially for discourse phenomena. 
  Popular metrics like BLEU are not expressive or sensitive enough to capture quality improvements or drops that are minor in size but significant in perception. We introduce the first of their kind MT benchmark datasets that aim to track and hail improvements across four main discourse phenomena: anaphora, lexical consistency, coherence and readability, and discourse connective translation. We also introduce evaluation methods for these tasks, and evaluate several baseline MT systems on the curated datasets. {Surprisingly,} we find that {existing} context-aware models do not improve discourse-related translations consistently across languages and phenomena. %We aim to maintain a leaderboard for future MT systems to demonstrate their competence at translating discourse phenomena. 
\end{abstract}

\section{Introduction and Related Work}
\label{sec:intro}

The advances in neural machine translation (NMT) systems have led to great achievements in terms of state-of-the-art performance in automatic translation tasks. There have even been claims that their translations are no worse than what an average bilingual human may produce \cite{Wu2016GooglesNM} or even that the translations are on par with professional  translators \cite{Hassan2018AchievingHP}. 

However, these claims only hold under a narrow set of controlled circumstances. When translations are evaluated monolingually or at the document level, human translations are preferred over MT outputs. \citet{Lubli2018HasMT} conduct extensive experiments for Chinese-English translations with professional translators, and find that although there is no statistical difference in adequacy between human and MT output at a sentence level, there is a statistically strong preference for human translations both in terms of adequacy and fluency when evaluated at the document level. Crucially, the document {(or discourse)} level phenomena {({\em e.g.,} coreference, coherence)} may not seem lexically significant but contribute significantly to readability and understandability of  {the translated} texts \blue{\cite{Guillou2012ImprovingPT}}.

{Meanwhile, there have been numerous attempts to model
extra sentential context for MT -- previously within the statistical MT \cite{carpuat-etal-2013-sensespotting,hardmeier-etal-2013-docent}, and recently  within the NMT framework. The NMT framework such as the Transformer \cite{NIPS2017_7181_vaswani} provides more flexibility to incorporate larger context. This has spurred a great deal of interest in developing context-aware NMT systems that take advantage of source or target contexts, \eg\ \cite{maruf-haffari-2018-document}, \cite{miculicich-etal-2018-document} and \cite{voita-etal-2018-context,voita-etal:2019b:ACL}, to name a few.}

%\cite{bawden-etal-2018-evaluating,voita-etal-2018-context,zhang-etal-2018-improving,miculicich-etal-2018-document,voita-etal:2019b:ACL,maruf-haffari-2018-document,maruf-etal-2019-selective}.} 

Despite the increasing interest in contextual MT, there is no framework for a principled comparison of results: there are no standard corpora and no agreed-upon evaluation measures. The selection of training datasets has mostly been arbitrary and much smaller in size compared to the standard ones (\eg\ WMT datasets).
%\footnote{{Annual Workshop (now Conference) on Machine Translation.}}

More critically, the lack of appropriate evaluation measures has been the key impediment in advancing contextual MT as it is important to measure if the context improves translations in terms of discourse phenomena, rather than mere improvements in lexical matching as is done with BLEU \cite{papineni2002bleu}. Indeed, recent studies also propose targeted datasets for evaluating phenomena like coreference \cite{guillou-etal-2014-parcor, guillou-hardmeier-2016-protest, lapshinova-koltunski-etal-2018-parcorfull,bawden-etal-2018-evaluating,voita-etal-2018-context}, and in the case of \cite{voita-etal:2019b:ACL}, testsets for ellipsis and lexical cohesion. The WMT-2019 tasks have also included document level translation and several adjoining user-submitted testsets targeted towards specific phenomena including subject-verb agreement, coreference, and others \cite{wmt-2018-machine-translation, wmt-2019-machine}. 

{In this work, we cover four diverse discourse phenomena using automatic data extraction methods, and also propose automatic evaluation methods for these tasks. Our targeted evaluation datasets are called the DiP benchmark tests (for \textbf{Di}scourse \textbf{P}henomena), that will allow us to compare models across discourse task strengths.} %, and also across languages. }

{Our analysis of state-of-the-art (SOTA) NMT models proves that testing a system on a single language pair is not sufficient as we observe significant differences in system  behavior and quality across languages. Our methods for automatically extracting testsets can be applied to multiple languages, and find cases that are difficult to translate without having to resort to synthetic data. 
Moreover, they can be easily updated to reflect current challenges, since datasets can become outdated as systems improve over the years.}

Our aim is to push the improvement of translation systems towards human-like output. Our main contributions in this paper are as follows:

\begin{itemize}[leftmargin=*]
\itemsep0em
    \item Benchmark datasets for four discourse phenomena: {anaphora}, {coherence \& readability}, {lexical consistency}, and {discourse connectives}.
    % \begin{itemize}
    %     \item Anaphora
    %     \item Coherence \& Readability
    %     \item Lexical Consistency
    %     \item Discourse Connectives
    % \end{itemize}
    \item Automatic evaluation methods and agreements with human judgments.
    \item Benchmark evaluation and analysis of three SOTA context-aware systems contrasted with baselines, for French/German/Russian-English language pairs.

    % Baseline results for Fr/De/Ru to En (and leaderboard)
   % \item \blue{Agreements with human judgments for some tasks}
      
\end{itemize}

{We open-source our framework at \href{https://ntunlpsg.github.io/project/discomt/DIP/}{https://ntunlpsg.github.io/project/discomt/DIP/}.}

\section{Machine Translation Models}
\label{sec:model}
We first introduce the baseline MT systems that we will be benchmarking in this work {and report their BLEU scores in our proposed setup.}

\subsection{Model Architectures}
We test the performance of three context-aware NMT models introduced by \citet{voita-etal-2018-context}, \citet{miculicich-etal-2018-document} and \citet{zhang-etal-2018-improving} on our DiP benchmark testsets.\footnote{{We excluded  \citet{maruf-haffari-2018-document,maruf-etal-2019-selective} as we found the implementation to be unoptimized and unable to train on a big dataset.}} Alongside, we also evaluate a sentence-level model, and a simple concatenation-based model \cite{tiedemann-scherrer-2017-neural} to contrast with the three elaborate context-aware models. 

% We also propose document-level data that contains discourse phenomena to train and {validate} the models.

% \vspace{-0.5em}
% \subsection{Model Architectures}

\begin{description}[leftmargin=0pt,itemsep=-0.1em]

\item [\sts: ] Our \sts\ baseline is a standard 6-layer base Transformer model \cite{NIPS2017_7181_vaswani} which translates sentences independently.

\item [\concat: ] Our \concat\ model is a 6-layer base Transformer whose input is two sentences (previous and current sentence) merged, with a special character serving as a separator.

\item [\anap:] \citet{voita-etal-2018-context} incorporate the source context by encoding it with a separate encoder, then fusing it in the last layer of a standard Transformer encoder using a gate. They claim that their model explicitly captures \emph{anaphora} resolution.  

\item [\han:] \citet{miculicich-etal-2018-document} introduce a \emph{hierarchical attention network} (\han) into the Transformer framework to dynamically attend to the context at two levels: word and sentence. % The authors evaluate the variations of \han\ by applying hierarchical attention separately to the encoder and decoder,
{They achieve} the highest BLEU when hierarchical attention is applied separately to both the encoder and the decoder. 

\item [\san:] \citet{zhang-etal-2018-improving} use a separate Transformer encoder to encode the context in the source side, which is then incorporated into the source encoder and target decoder using gates. We refer to this model as \emph{source attention network} (\san).

% as it uses the same  source-side context representation for both the encoder and decoder. 

For the context-aware models, we use the implementations from official author repositories. As the official code for \anap\ \cite{voita-etal-2018-context} has not been released, we implement the model in the Fairseq framework \cite{ott2019fairseq}.\footnote{\url{https://github.com/pytorch/fairseq}} {For training the \sts\ and \concat\ models we used the Transformer implementation from Fairseq.} 
We confirmed with the authors of \han\ and \san\ that our configurations were correct, and we took the best configuration directly from the \anap\ paper. Further details about the training settings and hyperparameters can be found in Appendix \ref{app:model_parameters}.

%\citet{zhang-etal-2018-improving} use a base Transformer model to compute the representation of document-level context, which  is  incorporated into both the encoder  and decoder using multi-head attention. 
%\vspace{-0.5em}
\end{description}

\subsection{Training Data}
%\red{To provide adequate data for models requiring one or more sentences of context (\eg\ \anap, \han, \san), we need sentence-aligned data with document boundaries.} Also, 

It is  essential to provide the models with training data that contains adequate amounts of discourse phenomena, if we expect them to learn such phenomena. To construct such datasets, we first manually investigated the standard WMT corpora consisting of UN \cite{ZIEMSKI16.1195}, Europarl \cite{TIEDEMANN12.463} and News Commentary, as well as the standard IWSLT dataset \cite{cettoloEtAl:EAMT2012}. We analyzed 100 randomly selected pairs of consecutive English sentences from each dataset, where the first sentence was treated as the context. Table \ref{table:datasets-disco-phenomena} shows the percentage of cases containing the respective discourse phenomena.

\begin{table}[t!]
%\footnotesize
%\tabcolsep=0.04cm
\setlength\extrarowheight{-2pt}
\scalebox{0.78}
{\begin{tabular}{l|cccc}
%\toprule
\textbf{Dataset} & \bf{Anaph.} & \bf{Lex. Con.} & \bf{Conn.} & \bf{ANY} \\
\midrule
{UN} & 0\% & 31\% & 0\% & 31\% \\
{Europarl} & 17\% & 24\% & 12\% & 49\% \\
{News Commentary} & 5\% & 18\% & 18\% & 37\% \\
{IWSLT} & 11\% & 19\% & 32\% & 42\% \\
\bottomrule
\end{tabular}
}
\vspace{-0.5em}
\caption{Discourse phenomena: \textbf{Anaph}ora {(restricted to anaphoric pronouns)}, \textbf{Lex}ical \textbf{Con}sistency, and Discourse \textbf{Conn}ectives in popular NMT datasets ({for English}). The column \textbf{ANY} shows the proportion of sentences which contain any of the listed phenomena. }
\label{table:datasets-disco-phenomena}
\end{table}

 In accordance with intuition, data sources based on narrative texts such as IWSLT exhibit increased amounts of discourse phenomena compared to strictly formal texts such as the UN corpus. On the other hand, the UN corpus consists of largely unrelated sentences, where only lexical consistency is well-represented due to the usage of very specific and strict naming of political concepts. 
%{Based on the finding that some datasets are better training material for discourse-aware systems than others, 
%and supported by additional experiments with model training on various dataset configurations using BLEU as a proxy quality measure, 
{We decided to {exclude the UN corpus} and combine the other datasets  {that have more discourse phenomena}. We evaluate the models} on the WMT-14 testset {which consists of news articles.} %is from news? [confirm]}. 
Table \ref{table:datasets_stats} shows the statistics of the resulting datasets. 

%We preprocess the datasets for three language pairs: Fr-En, De-En, and Ru-En.  Table \ref{table:datasets_stats}  shows the statistics. %The datasets and the preprocessing scripts will be released.  

\begin{table}[t!]
\centering
\setlength\extrarowheight{-5pt}
\scalebox{0.75}{
\begin{tabular}{lccc|c} 
\toprule
\textbf{Pair} & {\bf{Source}} & {\bf{Train}} & \bf{Dev} &  \bf{Test} \\
%             &               &               &          &  (WMT-14) \\
\midrule
\textbf{Fr-En} & IWSLT, Europarl, News  & 2,581,731  & 3,890  & 3,003 \\
\textbf{De-En} & IWSLT, Europarl, News & 2,490,871 & 3,693 & 3,003 \\
\textbf{Ru-En} & IWSLT, News & 459,572  & 4,777 & 3,003  \\
\bottomrule
\end{tabular}
}
\vspace{-0.5em}
\caption{{Dataset statistics for different language pairs in number of examples. The testset is from WMT-14.}}
\label{table:datasets_stats}
\end{table}

\subsection{BLEU Scores} 

The BLEU scores on the WMT-14 testset for each of the five trained models for De-En, Fr-En and Ru-En translation tasks are given in Table \ref{tab:bleu-scores}. 

We observe a variability in BLEU scores across the models.
{In contrast to increases in BLEU for selected language-pairs and datasets reported in the  published work, incorporating context within elaborate context-dependent models decreases BLEU scores for Fr-En and De-En.} \concat, the simple concatenation-based model,  achieves the best BLEU out of all of the tested models. This shows that context knowledge is indeed helpful for improving the BLEU.

For Ru-En task, dedicated context-aware models improve the performance. In particular, \anap\ achieves the highest score of all - interestingly, it has been trained and tested on En-Ru in the original paper \cite{voita-etal-2018-context}. This shows that complex architectures might only be useful for certain types of languages (such as highly inflected languages, like Russian).

\begin{table}[t!]
\centering
\setlength\extrarowheight{-5pt}
\scalebox{0.75}{
\begin{tabular}{lccc} 
\toprule
\textbf{Model} & {\bf{Fr-En}} & {\bf{De-En}} & \bf{Ru-En} \\
\midrule
\textbf{\sts} & 35.12 & 31.65 & 23.88 \\
\textbf{\concat} & \textbf{35.34} & \textbf{31.96} & 24.56  \\
\textbf{\anap} & 34.32 & 29.94 & \textbf{27.66}  \\
\textbf{\han} & 33.30 & 29.22 & 25.11 \\
\textbf{\san} & 33.48 & 29.32 & 26.24\\
\bottomrule
\end{tabular}
}
\vspace{-0.5em}
\caption{BLEU scores achieved by the benchmarked models on the WMT-14 testset.}
\label{tab:bleu-scores}
\end{table}

% \vspace{-0.3em}
% \subsection{Training Setup}
% %As a basis for our analysis of discourse aware state-of-the-art systems,

% \alert{Say breifly how the \sts\ and \concat\ were trained}

\section{Benchmark Testset Generation}
\label{sec:dataset}
% For the discourse phenomena that we propose to evaluate, : anaphora, coherence, lexical consistency, and discourse connectives,
We extract the testsets for {the} evaluated {discourse} phenomena automatically, based on existing errors in system outputs. This ensures that the data  {can \Ni provide hard contexts for translation without being artificial, \Nii be generated for multiple source languages, and \Niii be updated as frequently as possible; making them adaptable to errors in newer {(and possibly more accurate)} systems, and {making} the tasks harder over time.}
%keeping them up to date with the errors, making them adaptable to newer systems and also making the tasks harder over time. 

We use the system outputs released by WMT for the most recent years \cite{wmt-2017-machine, wmt-2018-machine-translation, wmt-2019-machine} to build our testsets. {For De-En, Fr-En and Ru-En, these consist of translation outputs from 51, 31 and 41 unique systems respectively. Since the data comes from a wide variety of systems, our testsets representatively aggregate different types of errors from several (arguably SOTA) models.}
%\red{For De, Fr and Ru this is 51, 31 and 41 unique systems respectively. is in fact representatively aggregated from several arguably SOTA models.}
{Also note that} the MT models we are benchmarking are not a part of these system submissions to WMT, so there is no potential bias in the testsets.

%Specifically, we align the system translations (English) to the references. Then, depending on the task, we automatically extract the sentences where the systems fail to translate correctly. The corresponding source texts of such translations form the benchmark testsets. Further refinement of the sentences is task dependent, the description of which follows. 

In this paper, we focus on translations from French, German, and Russian to English. We include French since Fr-En is a popular translation pair that results in some of the highest BLEU scores. WMT discontinued French from 2016 onwards, so the benchmark testsets for French  are smaller and based on relatively older 2013-2015 \cite{WMT2013, WMT2014, WMT2015} data. %; this overlaps with our validation data for training the benchmarked MT models.  
%Since the development and test set for French is consistent across all benchmarked MT systems, any improvements that may have resulted due to overlap in the fine-tuning data should be conferred across all systems, so we maintain that the results should still be comparable. 
{Other source languages that are part of WMT can be extracted as needed; the testsets can also be expanded if older data were to be considered}. The following sections describe the dataset, evaluation and verification procedures, and analysis of each of the discourse phenomena we benchmark.

\begin{comment}
In order to ensure that our methods of data extraction {(\ie\ the way we identify errors in translations)} and more particularly our evaluation methods align with human judgments, we conduct user studies for the tasks of anaphora, coherence and discourse connective {evaluations}. 
%\blue{Since the objective of our benchmarking task is to compare several systems against each other, we conduct studies requiring participants to judge the relative ranking of erroneous system translations for each of these tasks.--can remove} 
The participants were native/proficient bilingual speakers of English and were paid volunteers. Agreements for the studies are given in Table \ref{tab:study_agreements}.\footnote{Due to the nature of the dataset, the human annotators are more likely to choose the reference as the better candidate, which yields a skewed distribution of the annotations; traditional correlation measures such as Cohen's kappa are not robust to this, and thus we report the more appropriate Gwet's AC1/gamma coefficient \cite{gwetac1}.}
\end{comment}

\section{Anaphora}

{Anaphora are references to entities that occur elsewhere in a text; mishandling them can result in ungrammatical sentences or the reader inferring the wrong antecedent, leading to misunderstanding of the text \cite{Guillou2012ImprovingPT}. We focus specifically on the aspect of incorrect \emph{pronoun} translations.}
%Translating pronouns correctly is therefore necessary, and often requires contextual information.}

%GUILLOU:Protest
%\vspace{-0.5em}
\subsection{Pronoun Testset} 
%Anaphora has been the most studied discourse phenomenon in the context of discourse-based MT. 
To obtain hard contexts for pronoun translation, we look for source texts that lead to erroneous pronoun translations in recent WMT submissions.
We align the WMT system translations with their references, and collect the cases in which the translated pronouns do not match the reference. {This process requires the pronouns in the target language to be separate morphemes as in English.}  %\footnote{{Target language  must be separate morphemes.}} 

Our anaphora testset is an updated version of the one proposed by \citet{EvalPronom}, who also provide a list of cases where the translations can be considered wrong (rather than acceptable variants). We filter the system translations based on their list. The corresponding source texts are extracted as a test suite for pronoun translation. This gives us a pronoun benchmark testset with 1478 samples for Fr-en, 2245 samples for De-En and 2368 samples for Ru-En.

%\vspace{-0.5em}
\subsection{Pronoun Evaluation}

%\alert{Break the previous sentence into two.}
Targeted evaluation of pronouns in MT has been challenging as it is not fair to expect an exact match with the reference. 
%In the case of challenging testsets, there is the likelihood of getting very similar accuracies across MT outputs, as it is likely that all models will mistranslate the hard cases. 
Evaluation methods like APT \cite{APT} or AutoPRF \cite{Hardmeier2010ModellingPA} are specific to language pairs or lists of pronouns, requiring extensive manual intervention. {They have also been criticised} for failing to produce evaluations that are consistent with human judgments \cite{Liane-EMNLP18}.

\citet{EvalPronom} propose a model based evaluation measure for pronouns that generalizes well across language pairs and pronouns. They train a pairwise ranking model that scores ``good" pronoun translations (like in the reference) higher than the ``poor" pronoun translations (like in the MT output) {in context}, and show that their model is good at making this distinction, along with having high agreements with human judgements. {However, they do not rank multiple system translations against each other, {which is our main goal;} the absolute scores produced by their model are not useful since it is trained in a pairwise fashion.}

{We devise a way to use their model to score and rank system translations in terms of pronouns. First, we re-train their model with more up-to-date WMT data.\footnote{See Appendix {\ref{app:anaphora}} for details about the model training} We obtain a score for each benchmarked MT system (\sts, \concat, etc.) translation using the model, plus the corresponding reference sentence. We then \textit{normalize} the score for each translated sentence by calculating the difference with the reference. To get an overall score for an MT system, the assigned scores are summed across all sentences in the testset. }
\begin{equation}
\vspace{-0.4em}
\text{Score}_\text{sys} =  \sum_{i} \ps_{i}(\text{ref}|\theta) - \ps_{i}(\text{sys}|\theta) \label{eq:anap}
\vspace{-0.2em}
\end{equation}
\noindent where $\ps_{i}(\text{.}|\theta)$ denotes the score given to sentence $i$ by the pronoun model with parameters $\theta$. The systems are ranked based on this overall score, where a lower score indicates a better performance.

\begin{figure}
    \centering
    \includegraphics[scale=0.26]{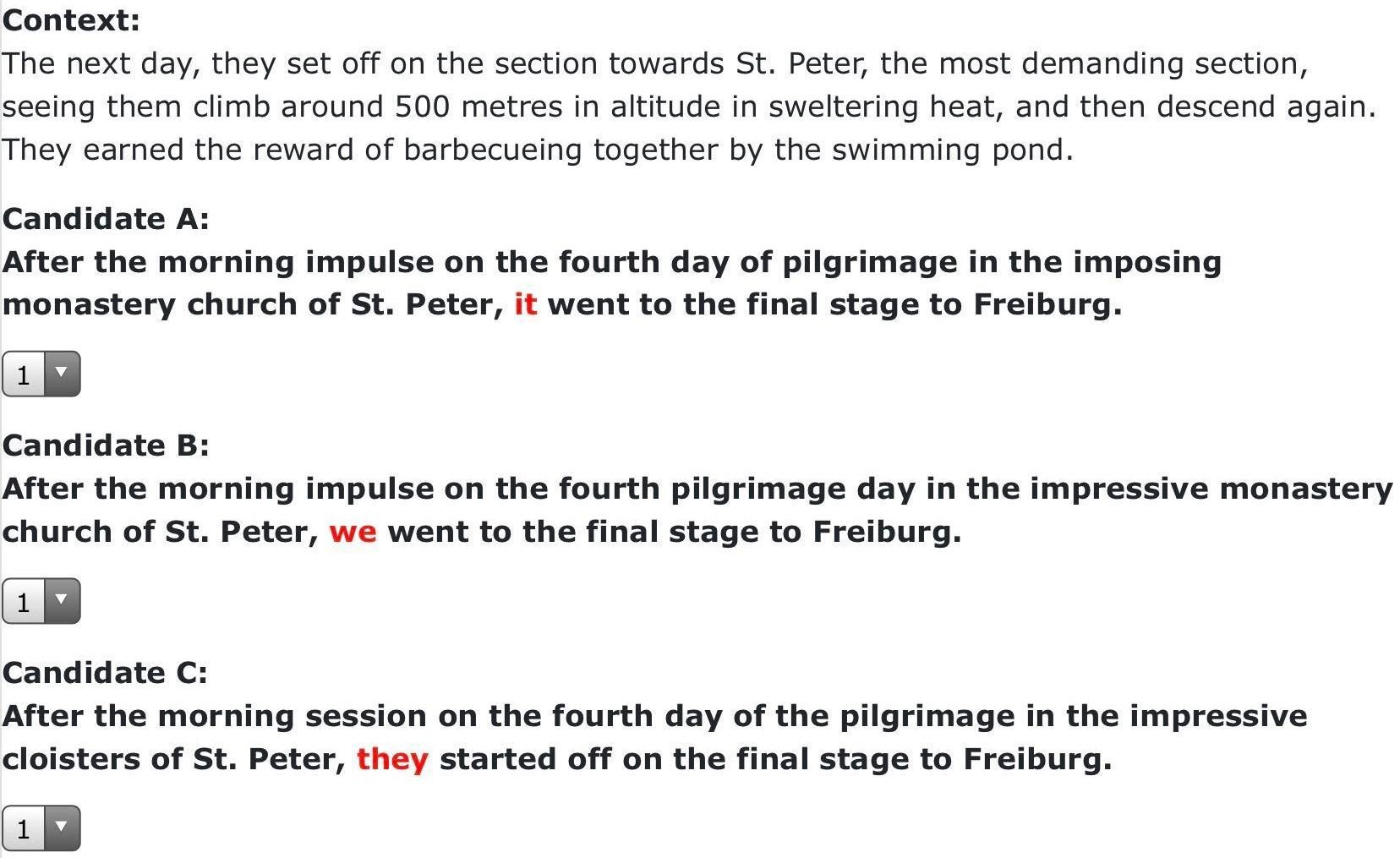}
    \vspace{-0.5em}
    \caption{User study interface (monolingual setup) for pronoun translation ranking. Pronouns in the sentences are highlighted in red.}
    \label{fig:prn_rank}
\end{figure}

%\vspace{-0.5em}
\paragraph{User study.}
{To confirm that our normalization-based ranking of systems agrees with human judgments, we conducted a user study.} {Participants are asked to rank given translation candidates in terms of their pronoun usage. We include the reference in the candidates, as a control.} %Participants were shown a source context of two sentences and the source sentence in bold, followed by three candidate translations of the source sentence. One of the candidates was the reference, as a control. The other two were translations with different pronoun errors produced by MT systems. See Appendix \ref{app:anaphora} for the user study interface.
We ask participants to rank system translations directly rather than a synthetically constructed contrastive pair (as was done by \citet{EvalPronom}) to ensure that our evaluations, which will be conducted on actual  translated texts, are reliable. %For a study that compares contrastive pairs, see \cite{EvalPronom}, who show that their model accurately focuses on pronouns. %\blue{}%See Fig.  for the study interface. 

{We first conducted the study in a \textit{bilingual} setup, in the presence of the source for German-English. Participants were shown a source context of two sentences and the source sentence in bold, followed by three candidate translations of the source sentence, one of which is the reference. The other two were translations with different pronoun errors produced by MT systems. Participants annotate 100 such samples. See Appendix \ref{app:anaphora} for the user study interface. }

We then conducted the study in a \emph{monolingual} setup without the source, \ie\ native speakers are shown the reference context in English, and the two candidate English translations and the reference translation as possible options for the sentence that follows {(Figure \ref{fig:prn_rank})}. To facilitate comparison, the data used for the German-English and only-English studies is the same.

The results are analysed to check \Ni how often the \emph{reference is preferred over the system translations} (our control), and \Nii how often the users agree in \emph{preference over the system translations} (\ie\ human judgment for translation quality). There were two participants in the bilingual setup, with the control experiment yielding an agreement of \textbf{0.72} {according to Gwet's AC1 \cite{gwetac1}}.\footnote{Due to the nature of the dataset, annotators are more likely to choose the reference as the better candidate, which yields a skewed distribution of the annotations; traditional correlation measures such as Cohen's kappa are not robust to this, and thus for this and all subsequent studies, we report the more appropriate Gwet's AC1/gamma coefficient. It is also the agreement reported by \cite{EvalPronom}.}  There were four participants in the monolingual setup, with the control yielding an {AC1} agreement of \textbf{0.82}{, which is higher than the bilingual setup.} We therefore use {the monolingual setup} to evaluate the rankings obtained from our modified evaluation method. We obtain an agreement of \textbf{0.91}, justifying the use of our {modified pronoun} model for evaluation.

%We therefore use the judgments from this study to evaluate the rankings obtained from our modified evaluation method.  

\subsection{Results and Analysis}

The ranking results obtained from evaluating the MT systems on our pronoun benchmark testset using our evaluation measure are given in Table~\ref{tab:anaphora_results_errors}. {We also report common pronoun errors for each model based on  our manual analysis.}

%\blue{ the findings from our manual analysis of pronoun translation errors.}

%Table~\ref{tab:anaphora_results_ranks} and 

{Overall,} we observe that surprisingly, \sts\ is translating pronouns comparatively well - outperforming all other models in De-En and Fr-En, and only giving way to \anap\ in Ru-En. %However, for De-En, \anap\ proved to be the worst contender so there is no clear tendency for this system's performance. \san\ is consistently poor in this task. 
The success of the {\sts\ model}
%\vanilla Transformer
can be explained by its tendency to use \textit{it} as the default pronoun, which statistically appears most often due to the lack of grammatical gender in English. More variability in pronouns occurs in the outputs of the context-aware models, but this does not contribute to a greater success.

\begin{table}[t]
%    \small
    \centering
    \scalebox{0.80}
  { \begin{tabular}
    {c|c|c|c|c}
    \toprule
    \multicolumn{5}{c}{De-En} \\
    \toprule
    \textbf{Rank} &  \textbf{Model} & \textbf{Gen Cp} & \textbf{NE} & \textbf{Lang}\\
    \midrule
    1 & \sts & 63 & 25 & 12 \\
    2 & \concat & 55 & 33 & 11 \\
    3 & \han & 44 & 22 & 33 \\
    4 & \san & 27 & 27 & 46 \\
    5 & \anap & 42 & 17 & 41 \\
    \bottomrule
    
    \toprule
    \multicolumn{5}{c}{Fr-En} \\
    \toprule
    \textbf{Rank} &  \textbf{Model} & \textbf{Gen Cp} & \textbf{NE} & \textbf{Lang}\\
    \midrule
    1 & \sts & 0 & 67 & 33 \\
    2 & \anap & 50 & 50 & 0 \\
    3 & \concat & 42 & 14 & 44 \\
    4 & \san & 43 & 29 & 28 \\
    5 & \han & 50 & 0 & 50 \\
    \bottomrule
    
    \toprule
    \multicolumn{5}{c}{Ru-En} \\
    \toprule
    \textbf{Rank} &  \textbf{Model} & \textbf{Gen Cp} & \textbf{NE} & \textbf{Lang}\\
    \midrule
    1 & \anap & 29 & 46 & 25 \\
    2 & \sts & 37 & 37 & 26 \\
    3 & \han & 31 & 48 & 21 \\
    4 & \concat & 29 & 46 & 25 \\
    5 & \san & 32 & 44 & 24 \\
    \bottomrule

    \end{tabular}}

    \caption{Pronoun evaluation: \textbf{Rank}ings of the different models for each language pair, obtained by summing the evaluation score for each sample in the pronoun benchmark. Each set of rankings is followed by the results of the manual analysis on a subset of the translation data. The percentages for the following types of errors are reported: Anaphora - instances of \textbf{Gen}der \textbf{C}o\textbf{p}y, \textbf{N}amed \textbf{E}ntity and \textbf{Lang}uage specific errors.}
    
    \label{tab:anaphora_results_errors}
\end{table}

Specifically, we observed the following types of errors in our manual analysis on a subset of the translation data:

\begin{description}[leftmargin=0pt,itemsep=-0.1em]
\vspace{-0.5em}
\item [\Ni Gender copy.]  
%This is a tendency to carry over gender from gender marked pronouns inherent in Fr/De/Ru languages. 
Translating from Fr/De/Ru to En often requires the `flattening' of gendered pronouns to \textit{it}, since Fr/De/Ru assign gender to all nouns. In many cases the machine translated pronouns tend to {(mistakenly)} agree with the source language. For example, % the sentence\textit{Mir wurde \textbf{diese Wohnung} in Earls Court gezeigt, und \textbf{sie} hatte ...}
{\textit{\textbf{diese Wohnung} in Earls Court..., und \blue{\textbf{sie}} hatte...} is translated to : \textit{\textbf{apartment} in Earls Court, and \red{\textbf{she}} had...},} a version which upholds the female gender expressed in \textit{sie}, instead of translating it to \textit{it}. 
%Ideally, the English text should have the pronoun \textit{it} instead of \textit{she}. 
{This was the most common error, except for Ru-En, where Named Entity errors were slightly more prevalent.}

% \textbf{Neuter.} This is a tendency to replace male and female personal pronouns with a neutral 'it' or plural 'they'. 

\item[\Nii Named entity.] A particularly hard problem is to infer gender from a named entity, \eg\ %\textit{... \textbf{Lady Liberty} is stepping forward. \textbf{She} is meant to be carrying the torch of liberty ...} 
{\textit{\textbf{Lady Liberty}...\blue{\textbf{She}} is meant to...}}- \textbf{she} is wrongly translated to \red{\textbf{it}}. Such  examples demand higher inference abilities such as world knowledge {(\eg\ distinguish male/female names).}
%(i.e., which names are male/female names? Which nouns are necessarily male/female?).

\item[\Niii Language specific phenomena.] {Pronouns can be ambiguous in the source language.}
%These are the phenomena specific to a language, where ambiguous pronouns can create additional problems for the model. 
For example in German, the pronoun 
% \textit{ihr} can mean \textit{her}, \textit{their}, or \textit{your}, while
\textit{sie} can mean both \textit{she} and \textit{you}, depending on capitalization, sentence structure, and context. This type of error often appears in the context-aware models, while being relatively rare for the {\sts\ model}. %vanilla Transformer.
\end{description}

\section{Coherence and Readability}

%Coherence and readability convey the logical relationships between sentences and make the text understandable as a whole \cite{Hobbs1979CoherenceAC}. 
{\citet{Pitler2008RevisitingRA} define coherence as the ease with which a text can be understood, and view readability as an equivalent property that indicates whether it is well-written.}
% \basia{{\citet{Pitler2008RevisitingRA} view coherence and readability as equivalent properties indicating whether the text is well-written and easy to understand.}}
It has been shown that NMT systems generate more fluent sentences than their phrase-based counterparts \cite{Castilho2017IsNM}. However, when the output is evaluated at the  document-level, it has also been shown that it lacks coherence \cite{Lubli2018HasMT}.

\begin{comment}
\begin{table}[t]
%    \small
    \centering
    \setlength\extrarowheight{-6pt}
    \scalebox{0.75}
  { \begin{tabular}
    {l|c|c}
    \toprule
    \textbf{Measure} & \textbf{\# Annotators} & \textbf{AC1 Agr.} \\
    \midrule
    \multicolumn{3}{c}{\textbf{Pronoun Translation Ranking}}\\
    \midrule
    Reference preferred (Fr-En) & 2 & 0.38 \\
    Reference preferred (De-En) & 2 & 0.72 \\
   % \midrule
    Reference preferred (Only-En) & 4 & 0.82 \\
    Agr. with model ranking & 4 & 0.91 \\
    \midrule
    \multicolumn{3}{c}{\textbf{Coherence Ranking}} \\
    \midrule
    Reference preferred & 3 & 0.84 \\
    Agr. with model ranking & 3 & 0.82 \\
    \midrule
    \multicolumn{3}{c}{\textbf{Discourse Connectives}} \\
    \midrule
    %Reference preferred & 2 & 0.82 \\
    Reference preferred & 2 & 0.98 \\
    \bottomrule
    \end{tabular}}
    \vspace{-0.5em}
    \caption{{AC1 agreements in the user studies for when \textbf{reference} was \textbf{preferred} by users over system translations, and \textbf{agreements with model rankings}. For pronoun translation ranking, agreements with model rankings are compared only for the only-English data.}}%\red{need to explain more what 'Higher/Lower ranked translation ..' means, maybe an example}
    
    \label{tab:study_agreements}
\end{table}
\end{comment}

%\vspace{-0.5em}
\subsection{Coherence Testset}

Our coherence and readability benchmarking is conducted at the document level; we try to find documents that can be considered hard to translate. To do this, we use the coherence model recently proposed by \citet{unifiedcoherence}, that achieves state-of-the-art results in  most coherence assessment tasks.  {The model has a Siamese framework, trained in a pairwise ranking fashion with positive and negative documents. The network models both syntax and inter-sentence coherence relations, along with global topic structures.} %Details about the training settings are given in the Appendix. 

%\footnote{\label{more-details}see Appendix for more details about the training.} 

% following the setup suggested by the authors
%We use the source code from {\href{https://ntunlpsg.github.io/project/coherence/n-coh-emnlp19/}{https://ntunlpsg.github.io/project/coherence/n-coh-emnlp19/}}; 

%Based on the coherence scores obtained by the different system translations for different texts,

The coherence model is originally trained on WSJ articles, where a negative document is formed by shuffling sentences of an original (positive) document. It needed to be re-trained with MT data to better capture the coherence issues that are present in MT outputs. {It has been shown in some studies that MT outputs are incoherent \cite{Smith2015APF,smith-etal-2016-trouble,Lubli2018HasMT}. We thus re-train the coherence model with reference translations as positive and MT outputs as negative documents}. We use the older WMT submissions from 2011-2015 for this re-training, to ensure that the training data does not overlap with the data used for extracting our benchmark testset. 

The model takes a system translation (multi-sentential) and its reference as input and produces a score for each. Similar to Eq. \ref{eq:anap}, we consider the difference between the scores produced by the model for the reference and the translated text as the coherence score for the translated text. 
 
%\red{We compute the mean coherence score for each unique source text $i$ as $\hat{c}_i &= \frac{1}{|\Ts_i|}\sum_{t \in \Ts_i} \cs_t$, where $\cs_t$ is the coherence score for translation $t$, and $\Ts_i$ is the set of system translations (\ie\ WMT submissions) for the source text $i$.} %($\hat{c}_i$)
{For a given source text (document) in the WMT testsets, we obtain the coherence scores for each of the translations (\ie WMT submissions) and average them. The source texts are then sorted based on the mean coherence scores of their translations. The texts that have lower mean coherence scores can be considered to have been hard to translate coherently. We threshold the scores to extract approximately the bottom 30\% of the texts {as a tradeoff between getting hard enough contexts and a reasonably-sized testset}. These \emph{source} texts  form our benchmark testset for coherence and readability. This yields 38 documents for Fr-En, 128 documents for De-En and 180 documents for Ru-En. }

%We normalize these scores across different system translations for each source text $i$. In particular, we   

%\begin{eqnarray}
%\vspace{-0.3em}
%c_i &= \frac{1}{|\Ts_i|}\sum_{t \in \Ts_i} \cs_t \label{eq:coh}%\\
%c_t &= s(\text{ref}|\theta) - s(\text{t}|\theta) 
%\vspace{-0.3em}
%\end{eqnarray}

%\vspace{-0.5em}
\subsection{Coherence Evaluation}
Coherence and readability is also a hard task to evaluate, as it can be quite subjective. We resort to model-based evaluation here as well, to capture the different aspects of coherence in translations.

%\footnote{\red{{\citet{smith-etal-2016-trouble} performed a similar evaluation with \textit{non-neural} coherence models and reported low accuracy.}}}
We use our {re-trained} coherence model to score the benchmarked MT system translations and modify the scores for use in the same way as the anaphora evaluation (Eq. \ref{eq:anap}) to obtain a relative ranking. As mentioned before (\cref{sec:dataset}), {the benchmarked MT systems do not overlap with the WMT system submissions, so there is no potential bias in evaluation since the testset extraction and the evaluation processes are independent.} To confirm that the model does in fact produce rankings that humans would agree with, and to validate our model re-training, we conduct a user study. 

%\blue{We use the coherence model scores for ranking in the same way as the anaphora evaluation: the differences between the scores of the references and the translations are used to rank the systems.} 

\vspace{-0.5em}
\paragraph{User study.}
\begin{figure}
    \centering
    \includegraphics[scale=0.26]{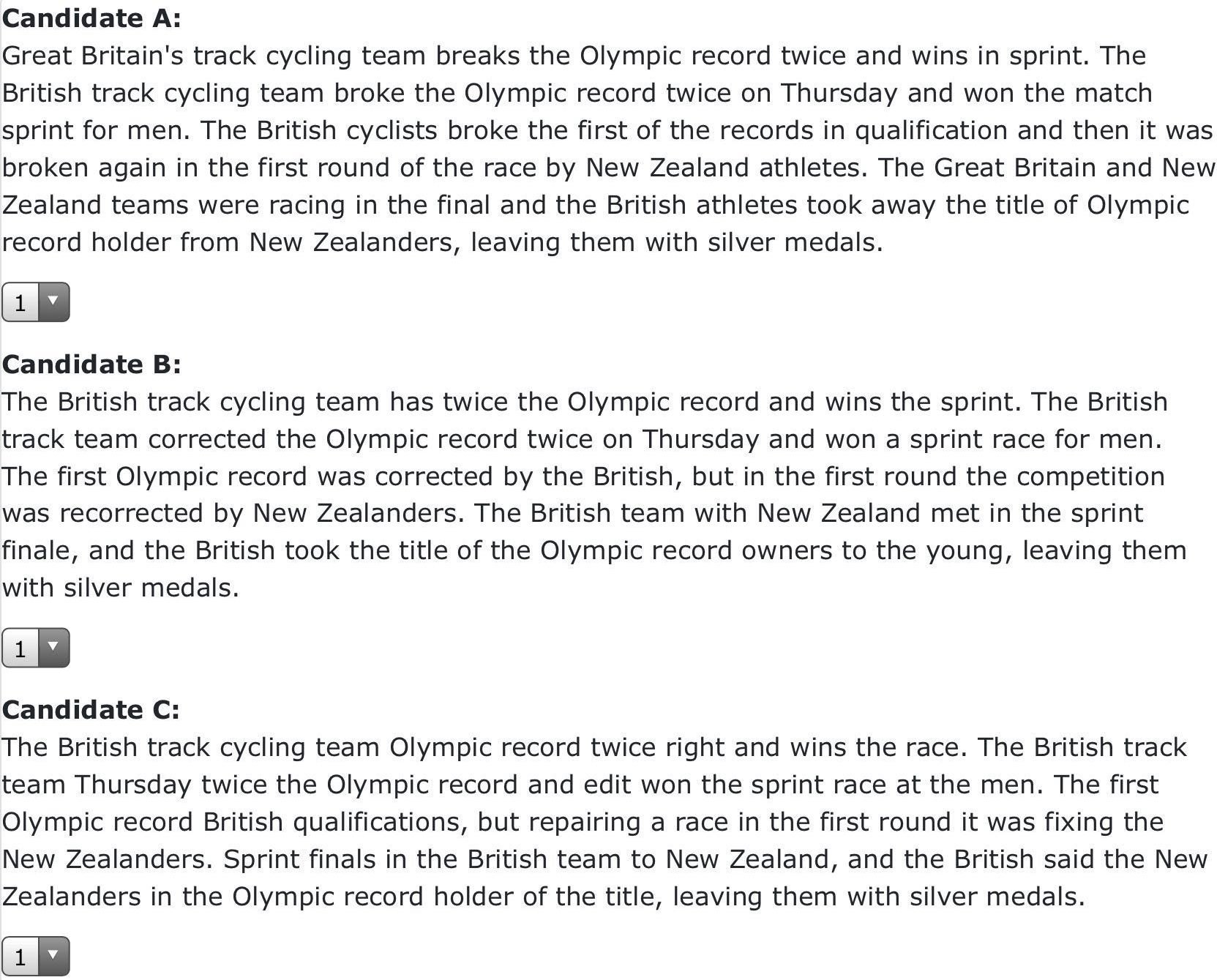}
    \caption{User interface for coherence study. The participants are shown 4-sentence texts and asked to rank them in terms of coherence and readability.}
    \label{fig:coh_rank}
\end{figure}
%We conduct a coherence ranking study for system translations. \footnote{See Appendix for the study interface.}
The participants are shown three candidate English translations of the same source text, and asked to rank the texts on how coherent and readable they are (Figure \ref{fig:coh_rank}). {To optimize annotation time, participants are only shown the first four sentences of the document}; they annotate 100 such samples. We also include the reference as one of the candidates for control, and to confirm that we are justified in re-training the evaluation model to assign a higher score to the reference.  %shows our user study interface.
%The user study interface is shown in the Appendix.

Three participants took part in the study. Our control experiment results in an AC1 agreement of \textbf{0.84}. The agreement between the human judgements and the coherence evaluation model's rankings is \textbf{0.82}. {The high agreement validates our proposal to use the modified coherence model to evaluate the benchmarked MT systems.}

%From Table \ref{tab:study_agreements}, we {see that the annotators' preference for the reference translation is high in this case as well, confirming our assumptions; the agreements between the annotators and the model's ranking are also high, enabling us to use the model to evaluate system translations}. 
%the participants preferred the reference 86.3\% of time. 

\subsection{Results and Analysis}
%The results of evaluating the MT systems on our coherence benchmark testset are given in Table~\ref{tab:coherence_results_ranks}. 
From the rankings in Table \ref{tab:coherence_results_ranks}, we see that \sts\ is the most coherent model for De-En and Ru-En. For Fr-En however, we observe an advantage of the context-aware model - \san, which ranks high for De-En as well. %The example of \san\ shows that a model can deal relatively well with one aspect of discourse translation (coherence) and do poorly in another (anaphora).
We identified the following types of coherence and readability errors (more examples in Appendix \ref{app:error_examples}).

%\basia{Manual qualitative analysis allows us to identify core reasons for loss of coherence. In Appendix we present examples of the errors.}

\begin{description}[leftmargin=0pt,itemsep=-0.1em]
\vspace{-0.5em}

\item [\Ni Inconsistency.] As in
\citet{somasundaran-etal-2014-lexical}, we observe that inconsistent translation of words across sentences (in particular named entities) breaks the continuity of meaning.

\item [\Nii Translation error.] Errors at various levels  spanning from ungrammatical fragments to model hallucinations introduce fragments which bear little relation to the whole text \cite{smith-etal-2016-trouble}. {An example of this:}

\blue{Reference}: \textit{There is huge applause for the Festival Orchestra, who appear on stage for the first time – in casual leisurewear in view of the high heat.} 

\red{Translation}: \textit{There is great applause for the solicitude orchestra , which is on the stage for the first time, with the heat once again in the wake of an empty leisure clothing.}

%\vspace{-0.7em}
\end{description}

% \red{Translation}: \textit{Farmers are allowed to start annual cage care on Monday.}
% \blue{Reference}: \textit{On Monday farmers are allowed to start the annual hedge bank maintainance.}

\begin{table}[t]
%    \small
    \centering
    \scalebox{0.85}
  { \begin{tabular}
    {c|c|c|c}
    \toprule
    
    \textbf{Rank} &  \textbf{De-En} & \textbf{Fr-En} & \textbf{Ru-En} \\
    \midrule
    1 & \sts & \san & \sts \\
    2 & \san & \sts & \anap \\
    3 & \concat & \concat & \concat \\
    4 & \anap & \anap & \san \\
    5 & \han & \han & \han \\
    \bottomrule

    \end{tabular}}

    \caption{Coherence {and Readability} evaluation: \textbf{Rank}ings of the different models for each language pair, obtained by summing evaluation scores for each document in the coherence benchmark testsets.}
    
    \label{tab:coherence_results_ranks}
\end{table}

\section{Lexical Consistency}
Lexical consistency in translation was first defined as `one translation per discourse' by \citet{Carpuat2009OneTP}, \ie\ the translation of a particular source word consistently to the same target word in that context. \citet{guillou-2013-analysing} analyze different human-generated texts and conclude that human translators tend to maintain lexical consistency, which supports the important elements in a text. {The consistent usage of lexical items in a discourse can be formalized by computing the \emph{lexical chains} \cite{Morris1991LexicalCC, LotfipourSaedi1997LexicalCA}.}

%\vspace{-0.5em}
\subsection{Lexical Consistency Testset}
% Lexical consistency in translation was first defined as `one translation per discourse' by \citet{Carpuat2009OneTP}, \ie\ the translation of a particular source word consistently to the same target word in that context. \citet{guillou-2013-analysing} analyze different human-generated texts and conclude that human translators tend to maintain lexical consistency, which supports the important elements in a text. 
{To extract a testset for lexical consistency evaluation,} we first align the translations from WMT submissions with their references. In order to get a reasonable lexical chain formed by a consistent translation, we consider translations of blocks of 3-5 sentences in which the (lemmatized) word we are considering occurs at least twice in the reference. For each such word, we check if the corresponding system translation produces the same (lemmatized) word at least once, but fewer than the number of times the word occurs in the reference. In such cases, the system translation has failed to be lexically consistent in translation (see Figure \ref{fig:lexcon_eg} for an example). We limit the errors considered to nouns and adjectives. The source texts of these cases form the benchmark testset. {This gives us a testset with 172 sets of sentences for Fr-En, 312 sets for De-En and 358 sets for Ru-En.}

\begin{figure}
\small
{
{\textbf{Reference translation}: \emph{You are still missing the \textbf{\blue{union}} of married men... I'll form a \textbf{\blue{union}}... protected professions equipped with \textbf{\blue{unions}} quite enough, which created... }}\\
%\vspace{1em}
{\textbf{System Translation}: \emph{They still lack the \textbf{\red{sindicat}} sdes married men... I am the french \textbf{\red{syndicate}}... protected occupations with trade \textbf{\red{unions}}, which created...}
}

}
\vspace{-0.3em}   
\caption{Lexical consistency maintained in the reference but not in a system translation (WMT-2015 Fr-En)} % \alert{use red/blue color?}}
\label{fig:lexcon_eg}
\normalsize
\end{figure}

One possible issue with this method could be that reference translations may contain forced consistency, \ie\ human translators introduce consistency to make the text more readable, despite inconsistent word usage in the source. It may not be reasonable to expect consistency in a system translation if there is none in the source. {To confirm, we conducted a manual analysis where we compared the lexical chains of nouns and adjectives in Russian and French source texts against the lexical chains in the English reference. We find that in a majority {(77\%)} of the cases, the lexical chains in the source are reflected accurately in the reference, and there are relatively few cases where humans force consistency. Considering the fact that the same data is used for BLEU calculations, we presume that this should not be a significant issue.}

%\vspace{-0.5em}
\subsection{Lexical Consistency Evaluation}

For lexical consistency, we adopt a simple evaluation method. For each block of 3-5 sentences, we either have a consistent translation of the word in focus, or the translation is inconsistent. We simply count the instances of consistency and rank the systems based on the percentage of accuracy.

It is possible that the word used in the system translation is not the same as the word in the reference, but the MT output is still consistent (\eg\ a synonym used consistently). We tried to use alignments coupled with similarity obtained from ELMo \cite{PetersELMo:2018} and BERT \cite{devlin2018bert} embeddings to evaluate such cases to avoid unfairly penalizing the system translations, but we found this to be noisy and unreliable. Thus, we match with the reference, as it can be argued that such words are salient and therefore must be translated exactly to convey the correct meaning. %Since this is a specifically designed task without a model-based evaluation, we do not conduct a user study.

\begin{table}[t]
%    \small
    \centering
    \scalebox{0.78}
  { \begin{tabular}
    {c|c|c|c|c|c|c|c}
    \toprule
    \multicolumn{8}{c}{De-En} \\
    \toprule
    \textbf{Rk} & \textbf{Model} &  \textbf{Acc} & \textbf{Syn} & \textbf{Rel} & \textbf{Om} & \textbf{NE} & \textbf{Rd}\\
    \midrule
    1 & \concat & 42.30 & 38 & 15 & 23 & 4 & 19 \\
    2 & \anap & 38.14 & 46 & 21 & 21 & 4 & 8 \\
    3 & \sts & 36.85 & 38 & 19 & 29 & 5 & 9 \\
    4 & \han & 36.21 & 35 & 22 & 30 & 4 & 7 \\
    5 & \san & 35.57 & 38 & 19 & 24 & 5 & 14 \\
    \bottomrule
    
    \toprule
    \multicolumn{8}{c}{Fr-En} \\
    \toprule
     \textbf{Rk} &  \textbf{Model} & \textbf{Acc} & \textbf{Syn} & \textbf{Rel} & \textbf{Om} & \textbf{NE} & \textbf{Rd}\\
    \midrule
    1 & \han & 36.21 & 48 & 26 & 4 & 0 & 22 \\
    2 & \sts & 36.04 & 43 & 19 & 19 & 0 & 19 \\
    3 & \anap & 30.81 & 35 & 25 & 15 & 0 & 25 \\
    4 & \concat & 30.81 & 35 & 25 & 15 & 0 & 25 \\
    4 & \san & 30.81 & 44 & 12 & 12 & 0 & 32 \\
    
    \bottomrule
    
    \toprule
    \multicolumn{8}{c}{Ru-En} \\
    \toprule
     \textbf{Rk} &  \textbf{Model} & \textbf{Acc} & \textbf{Syn} & \textbf{Rel} & \textbf{Om} & \textbf{NE} & \textbf{Rd}\\
    \midrule
    1 & \anap & 13.68 & 15 & 0 & 26 & 15 & 44\\
    2 & \sts & 10.33 & 21 & 9 & 27 & 21 & 21 \\
    2 & \concat & 10.33 & 15 & 8 & 15 & 18 & 44 \\
    3 & \san & 8.37 & 6 & 9 & 24 & 18 & 42 \\
    4 & \han & 5.58 & 11 & 8 & 19 & 19 & 41 \\
    \bottomrule

    \end{tabular}}

    \caption{Lexical consistency evaluation: \textbf{R}an\textbf{k}ings of the different models for each language pair, ranked by their \textbf{Acc}uracy. Accuracy here is defined as the percentage of samples in the benchmark dataset translations in which the models maintain lexical consistency. Each set of rankings is followed by the results of the manual analysis on a subset of the translation data for \textbf{Syn}onyms, \textbf{Rel}ated words, \textbf{Om}issions, \textbf{N}amed \textbf{E}ntity, \textbf{R}an\textbf{d}om translation. }
    
    \label{tab:lexcon_results_errors}
\end{table}

\subsection{Results and Analysis}

The rankings of the MT systems based on accuracy on the lexical consistency benchmark testsets are given in Table~\ref{tab:lexcon_results_errors}, along with our findings from a manual analysis on a subset of the translations. %Table~\ref{tab:lexcon_results_ranks} and 

{The overall low quality of Russian translations contributes to the prevalence of \textbf{R}an\textbf{d}om translations, and the necessity to transliterate named entities increases \textbf{NE} errors, compared to other languages.}
 \concat\ and  \sts\ are again successful {on average, taking the first or second place in all tested languages}, while {\anap\ leads the board again for Ru-En}. Our manual inspection of the lexical chains shows the following tendencies:

\begin{description}[leftmargin=0pt,itemsep=-0.2em]
\vspace{-0.5em}
\item [\Ni Synonym \& related word.] Words are exchanged for their synonyms (\textit{poll} - \textit{survey}), hypernyms/hyponyms  (\textit{ambulance} - \textit{car}) or related concepts (\textit{wine} - \textit{vineyard}).

%For lexical consistency, synonymous translations are not welcome since we expect all words in a lexical chain to be translated exactly the same.

\item [\Nii Named entity.] Models tend to distort proper names and translate them inconsistently. For example, the original name  \textit{Füchtorf} (name of a town) gets translated to \textit{feeding-community}.

\item [\Niii Omission.] Occurs when words are omitted altogether from the lexical chain.

%\vspace{-0.7em}
\end{description}

%taking the first and second place in De-En and Fr-En, respectively.
\begin{comment}

\begin{table}[t]
%    \small
    \centering
    \scalebox{0.80}
  { \begin{tabular}
    {c|c|c|c}
    \toprule
    
    \textbf{Rk} &  \textbf{De-En} & \textbf{Fr-En} & \textbf{Ru-En} \\
    \midrule
    1 & \concat & \han & \anap \\
    2 & \anap & \sts & \sts, \concat \\
    3 & \sts & \anap & \san \\
    4 & \han & \concat, \san & \han \\
    5 & \san & -  & - \\
    \bottomrule

    \end{tabular}}

    \caption{Lexical Consistency testset: \textbf{R}an\textbf{k}ings of the different models for each language pair, based on the accuracy. Accuracy here is defined as the percentage of samples in the benchmark dataset translations in which the models maintain lexical consistency.}
    
    \label{tab:lexcon_results_ranks}
\end{table}
\end{comment}

\section{Discourse Connectives}

{Discourse connectives are used to link the contents of texts together by signaling coherence relations that are essential to the understanding of the texts \cite{PDTB:Reflections}. Failing to translate a discourse connective correctly can result in texts that are hard to understand or ungrammatical.}

%\vspace{-0.5em}
\subsection{Discourse Connective Testset}
Finding errors in discourse connective translations can be quite tricky, since there are often many acceptable variants. %\red{Studies have shown inconclusive results when users were asked to choose the correct connective from a list for a given context \cite{malmi-etal-2018-automatic}; many conducted studies have contradicting results \cite{Geva2019DiscoFuseAL, Patterson2013PredictingTP}.} 
To mitigate confusion, we limit the errors we consider in discourse connectives to the setting where the reference contains a connective but the translations fail to produce any. 

Although there is an accepted list of explicit discourse connectives, it would not be appropriate to simply extract such cases, since those words may not always act in the capacity of a discourse connective. In order to identify the discourse connectives, we build a simple explicit connective classifier (a neural model) using annotated data from the Penn Discourse Treebank or PDTB \cite{PDTBv3}. {The classifier achieves an average cross-validation F1 score of \textbf{93.92} across the 25 sections of PDTBv3, proving that it generalizes well.} {See Appendix \ref{app:connectives} for more details about the model.}

%The model consists of an LSTM layer \cite{hochreiter1997long} followed by a linear layer for binary classification, initialized by ELMo embeddings \cite{Peters:2018} and achieves P/R/F1 scores of 96.01/93.08/94.53 respectively.  
After identifying the explicit connectives in the reference and the system translations, we align them and extract the source texts of cases with missing connective translations. {We only use the classifier on the reference text, but consider all possible markers in the system translations to give them the benefit of the doubt.} {This gives us a discourse connective benchmark testset with 109 samples for Fr-En, 109 samples for De-En and 117 samples for Ru-En.} %See {Appendix} for details about the connective classifier and the list of connectives represented in the dataset.

%We give details about the model in the Appendix.
%\vspace{-0.5em}
\subsection{Discourse Connective Evaluation}
{There has been some work on semi-automatic evaluation of translated discourse connectives in \citet{Meyer2012MachineTO} and \citet{Hajlaoui2013AssessingTA}; however, it is limited to only En-Fr, based on a dictionary list of equivalent connectives, and requires using potentially noisy alignments and other heuristics.}  
%Due to inconsistent results from studies as mentioned before, 
In the interest of evaluation simplicity, we expect the model to produce the same connective as the reference. Since the nature of the challenge is that connectives tend to be omitted altogether, we report both the accuracy of connective translations with respect to the reference, and the percentage of cases where \textit{any} candidate connective is produced.  

%\red{unclear sentence: } It can be argued that the reference translation  accurately reflects the specific relation that the connective conveys. Therefore, it may not be unreasonable to expect a matching connective in the translation. 
\begin{figure}
    \centering
    \includegraphics[scale=0.29]{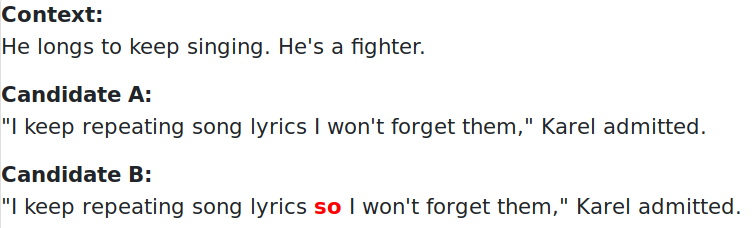}
    %\vspace{-0.5em}
    \caption{{Connective study interface. Participants are shown the reference with the connective and another option without the connective, and asked to choose the best option that follows the given context.}}
    \label{fig:conn_study}
\end{figure}

\vspace{-0.5em}
\paragraph{User study.}
To confirm that the presence of the connective conveys some information and contributes to better readability and understanding of the text, we conduct a user study. As presented in Figure \ref{fig:conn_study}, participants are shown two previous sentences from the reference for context, and asked to choose between two candidate options for the sentence that may follow. These options consist of {the reference translation with the connective highlighted, and the same text with the connective deleted. {{We also conducted a study using system translations with missing connectives directly; see Appendix \ref{app:connectives} for discussion}}}. %a system translation that is missing a translated connective.

Participants are asked to choose the sentence which more accurately conveys the intended meaning. There were two participants who annotated 200 such samples. The reference with the connective was chosen over the version without the connective with an AC1 agreement of \textbf{0.98}. {See Appendix \ref{app:connectives} for connective-wise results.} {Note that participants may prefer the version with the connective due to loss of grammaticality or loss of sense information when the connective is missing. Although indistinguishable in this setting, we argue that since both affect translation quality, it is reasonable to expect a translation for the connectives.}  %See Appendix for a connective-wise breakdown of the results. }

\subsection{Results and Analysis}

The rankings of MT systems based on their accuracy of connective translations are given in Table~\ref{tab:discon_results_errors}, along with our findings from a manual analysis on a subset of the translations. The ranking shows that \sts\ models are on average the most accurate and {omit the connectives less often}. {\anap\ continues its high performance in Ru-En, and while \san\ leads the board for De-En in terms of accuracy, it has a low percentage of cases overall in which any connective is produced.}  %\han\ models had a clear tendency to omit connectives and were found to be consistently poor in this task. Other models show no clear tendency. 

In benchmark outputs, we observed mostly \textbf{omissions} of connectives (disappears in the translation), \textbf{synonymous translations} (\eg\  \textit{Naldo is \blue{\textbf{also}} a great athlete on the bench} - \textit{Naldo's ``great sport" on the bank, \red{\textbf{too}}.}), and \textbf{mistranslations}.

\begin{comment}
\begin{table}[t]
%    \small
    \centering
    \scalebox{0.85}
  { \begin{tabular}
    {c|c|c|c}
    \toprule
    
    \textbf{Rank} &  \textbf{De-En} & \textbf{Fr-En} & \textbf{Ru-En} \\
    \midrule
    1 & \san & \concat & \sts \\
    2 & \sts & \sts & \anap \\
    3 & \anap & \han & \san \\
    4 & \concat & \san & \concat \\
    5 & \han & \anap  & \han \\
    \bottomrule

    \end{tabular}}

    \caption{Discourse Connective testset: \textbf{Rank}ings of the different models for each language pair, first by accuracy of connective translation, then by the percentage of cases where any candidate connective was produced.}
    
    \label{tab:discon_results_ranks}
\end{table}
\end{comment}

\begin{table}[t]
%    \small
    \centering
    \scalebox{0.80}
  { \begin{tabular}
    {c|c|c|c|c|c|c}
    \toprule
    \multicolumn{7}{c}{De-En} \\
    \toprule
    \textbf{Rank} & \textbf{Model} &  \textbf{Acc} & \textbf{ANY} & \textbf{Om} & \textbf{Syn} & \textbf{Mis}\\
    \midrule
    1 & \san & 52.29 & 76.15 & 67 & 33 & 0 \\
    2 & \sts & 50.46 & 78.90 & 76 & 24 & 0 \\
    3 & \anap & 50.46 & 76.5 & 75 & 25 & 0 \\
    4 & \concat & 46.79 & 75.23 & 68 & 32 & 0 \\
    5 & \han & 46.79 & 67.89 & 72 & 28 & 0 \\
    \bottomrule
    
    \toprule
    \multicolumn{7}{c}{Fr-En} \\
    \toprule
    \textbf{Rank} & \textbf{Model} &  \textbf{Acc} & \textbf{ANY} & \textbf{Om} & \textbf{Syn} & \textbf{Mis}\\
    \midrule
    1 & \concat & 48.62 & 76.15 & 47 & 50 & 3 \\
    2 & \sts & 48.62 & 75.23 & 53 & 44 & 2 \\
    3 & \han & 46.79 & 71.56 & 53 & 43 & 3 \\
    4 & \san & 46.79 & 70.64 & 56 & 41 & 2  \\
    5 & \anap & 46.79 & 68.81 & 53 & 41 & 6 \\
    
    \bottomrule
    
    \toprule
    \multicolumn{7}{c}{Ru-En} \\
    \toprule
    \textbf{Rank} & \textbf{Model} &  \textbf{Acc} & \textbf{ANY} & \textbf{Om} & \textbf{Syn} & \textbf{Mis}\\ 
    \midrule
    1 & \sts & 40.17 & 76.92 & 59 & 28 & 12\\
    2 & \anap & 39.32 & 68.38 & 63 & 30 & 7 \\
    3 & \san & 39.32 & 64.96 & 62 & 28 & 9 \\
    4 & \concat & 35.04 & 75.08 & 61 & 32 & 6 \\
    5 & \han & 33.34 & 57.26 & 76 & 21 & 3 \\
    \bottomrule

    \end{tabular}}

    \caption{Discourse connective evaluation: \textbf{Rank}ings of the different models for each language pair, ranked first by their \textbf{Acc}uracy and then by the percentage where \textbf{ANY} connective is produced. Each set of rankings is followed by the results of the manual analysis on a subset of the translation data for \textbf{Om}issions, \textbf{Syn}onyms, \textbf{Mis}translations. }
    
    \label{tab:discon_results_errors}
\end{table}

% \begin{table}[t]
%     \small
%     \centering
%     \begin{tabular}{c|c}
%     \toprule
%     \textbf{Measure} & \textbf{AC1 Agr.}  \\
%     \midrule
%     Reference preferred & 0.82 \\
%     %Agr. with Model ranking & 0.83  \\
%   % Lower ranked translation by the model & 0.86 & 87.0 \\
%     \end{tabular}
%     \caption{AC1 agreements between the participants on whether they preferred the reference with the connective or the system translation without the connective. No. of annotators = 2.}
%     \label{tab:conn_study_agreements}
% \end{table}

\begin{comment}
\begin{table}
\centering
\setlength\extrarowheight{-6pt}
\scalebox{0.85}{\begin{tabular}{l|c|c|c}  
%\toprule
\textbf{Tasks} & \bf{Fr-En} & \bf{De-En} & \bf{Ru-En} \\
\midrule
Anaphora & 1478 & 2245 & 2368 \\
Lexical Consistency & 172 & 312 & 358 \\
Coherence & 38 & 128 & 180 \\
Discourse Connectives & 109 & 109 & 117 \\
\bottomrule
\end{tabular}
}
\vspace{-0.5em}
\caption{Testset statistics (number of samples). For anaphora and discourse connectives, each sample consists of one sentence with two previous sentences for context. For coherence, each sample is a full document. For lexical consistency, each sample is a set of 3-5 sentences that contains a lexical chain.}
\label{table:testsets}
\end{table}
\end{comment}

%\section{Results and Analysis}
%\label{sec:eval}
%\input{evaluation.tex}

\section{Discussion}
\label{sec:discussion}
{Our benchmark evaluation on various discourse phenomena across different MT systems and language pairs reveals gaps in evaluation results that are typically reported. A lack of comprehensive evaluation makes it difficult to determine which models perform conclusively better than others.}

{Our results re-emphasize the gap between BLEU scores and translation quality at the discourse level. The overall BLEU scores for Fr-En are higher than the BLEU scores for De-En; however, we see that both the lexical consistency and the discourse connective accuracies are higher for De-En. %This makes it uncertain whether the quality of translations for Fr-En is indeed better than those for De-En. 
{Similarly, for Ru-En, both \san\ and \han\ have higher BLEU scores than the \sts\ and \concat\ models, but are unable to outperform these simpler models consistently in the discourse tasks, often ranking last.}}

{We also reveal a gap in performance consistency across language pairs. Models may be tuned for a particular language pair, such as \anap\, which was trained for En-Ru. For the same language pair (Ru-En), we show results consistent with what is reported; the model leads the board for anaphora and lexical consistency, while ranking second for coherence and readability, and discourse connectives. {However, it is not so successful in other languages, ranking at the bottom for anaphora in De-En and discourse connectives in Fr-En, and close to bottom for coherence in Fr-En and De-En.} \san\ performs highly in coherence for Fr-En and De-En, in contrast to its performance on other tasks and languages; the authors originally report improved results for Fr-En.  }

{In general, our findings match the conclusions from \citet{context-study} regarding the lack of satisfactory performance gains in context-aware models. Given no comprehensive evaluation across language pairs, the best bet for training an MT model is to use the baseline \sts\ and \concat\ models, which perform more or less reliably across different tasks. Our results emphasize the need for standard benchmarking datasets and evaluation measures across language pairs, that will provide a better picture of MT system performance. } 

{Although some of the testsets we provide are limited in size, it is a consequence of favouring precision to maintain data quality and limiting data to recent years. However, since the extraction is automatic, the datasets can be extended as submissions are added {to the upcoming evaluation campaigns}, while also increasing the difficulty of the tasks as MT systems improve. We hope that the discourse benchmark testsets and evaluation procedures we provide can contribute towards a more comprehensive MT evaluation framework, and prove useful in obtaining a more complete idea of a system's translation quality.  }

%{In general, our findings match the conclusions from \citet{context-study} regarding the lack of satisfactory performance gains in context-aware models. However, our results stress that the problem is not caused by context being unhelpful or unrepresented by BLEU, but rather by the faulty behavior of the models themselves, as the simple contextual model \concat\ is able to beat both context-aware models as well as a sentence-level model.}

\section{Conclusions}
\label{sec:conc}
We presented the first of their kind discourse phenomena based benchmarking testsets called the DiP tests, designed to be challenging for NMT systems. %We also present automatic evaluation methods that agree with human judgments. We evaluate several NMT models on our testsets and provide results and extensive analyses of their performance for three source languages. Our testsets and model-based evaluations can be automatically updated to increase difficulty as systems improve through the years. 
We show that complex context-aware models are not consistent in their performance. %, and also that BLEU scores correspond poorly to translation performance on discourse phenomena. 
Our main goal is to motivate the benchmarking of MT systems with more indicative performance yardsticks. We will release the document-level training corpora and discourse benchmark testsets for public use, and also propose to accept translations from MT systems to maintain a leaderboard for the described phenomena.

%In future work, we hope to widen the range of phenomena covered in the test sets, and to develop more sophisticated evaluation methods that are fine-grained and lenient towards varied but equivalent correct translations. 

%\section{Model - Baselines}
%\label{sec:model}
%\input{model.tex}

%\section{Results}
%\label{sec:results}
%\input{results.tex}

%\section{Conclusions and Future Work}
%\label{sec:conclusions}
%\input{conclusions.tex}

\bibliography{bmark}
\bibliographystyle{acl_natbib}

\appendix
\section{Appendix}
\label{sec:appendix}
%\appendix
%\label{sec:appendix}
%\section{Appendix}

\subsection{Anaphora}
\label{app:anaphora}

\paragraph{Re-trained model.}
The pronoun evaluation model results reported in \citet{EvalPronom} is based on a model that is trained on WMT11-15 data and tested on WMT-2017 data. We re-train the model with more up-to-date data from WMT13-18, and test the model on WMT-19 data. {Note that this training data is taken from WMT submissions, which do not overlap with the benchmarked MT models; there is therefore no conflict in using this trained model to evaluate the benchmarked model translations.} Results are shown in Table \ref{results:pronoun-model}. Their model scores the translations in context; we provide the previous two sentences from the reference translation as context according to their settings. %Figure \ref{fig:prn_rank} shows our user study interface.

\begin{figure}
\small
{\textbf{System translation 1}: \emph{Biles became the fourth consecutive American \textbf{woman }to win the title of absolute champion of the Championship and the fifth in General, consolidating\textbf{ its} reputation as the best of \textbf{his} generation and maybe ever. } 

\vspace{0.2em}

\textbf{System translation 2}: \emph{Biles became the fourth consecutive American \textbf{woman} to win the title of champion absolute Championship and fifth in general, perpetuating \textbf{his} reputation as the best of \textbf{his} generation and perhaps ever.}
}
\vspace{-0.3em}
\caption{Example taken from \cite{EvalPronom}; two system translations of the same source, with different pronoun errors (correct: her and her).} %The model from \cite{EvalPronom} scores the second one higher; the animacy and consistent pronoun translation are claimed to be contributing factors.}
\normalsize
\label{fig:prn_rank_example}
\end{figure}

% \begin{figure*}
%     \centering
%     \includegraphics[scale=0.25]{figures/german_pronoun_rank_3.png}
%     \caption{User Study Interface for Pronoun ranking}
%     \label{fig:prn_rank}
% \end{figure*}

\begin{table}[h]
\centering
\small
\tabcolsep=0.1cm
\begin{tabular}{l|c|c}  
\toprule
\textbf{Training data} & \textbf{Test data}  &  \textbf{Accuracy} \\
\midrule
WMT13-18 & WMT-19 & 86.76 \\
\bottomrule
\end{tabular}
\caption{Results of the re-trained pronoun scoring model.}
\label{results:pronoun-model}
\end{table}

\paragraph{Evaluation example.}
An example comparing different pronoun errors against each other from \citet{EvalPronom} is in Figure \ref{fig:prn_rank_example}.

\begin{figure}
    \centering
    \includegraphics[scale=0.26]{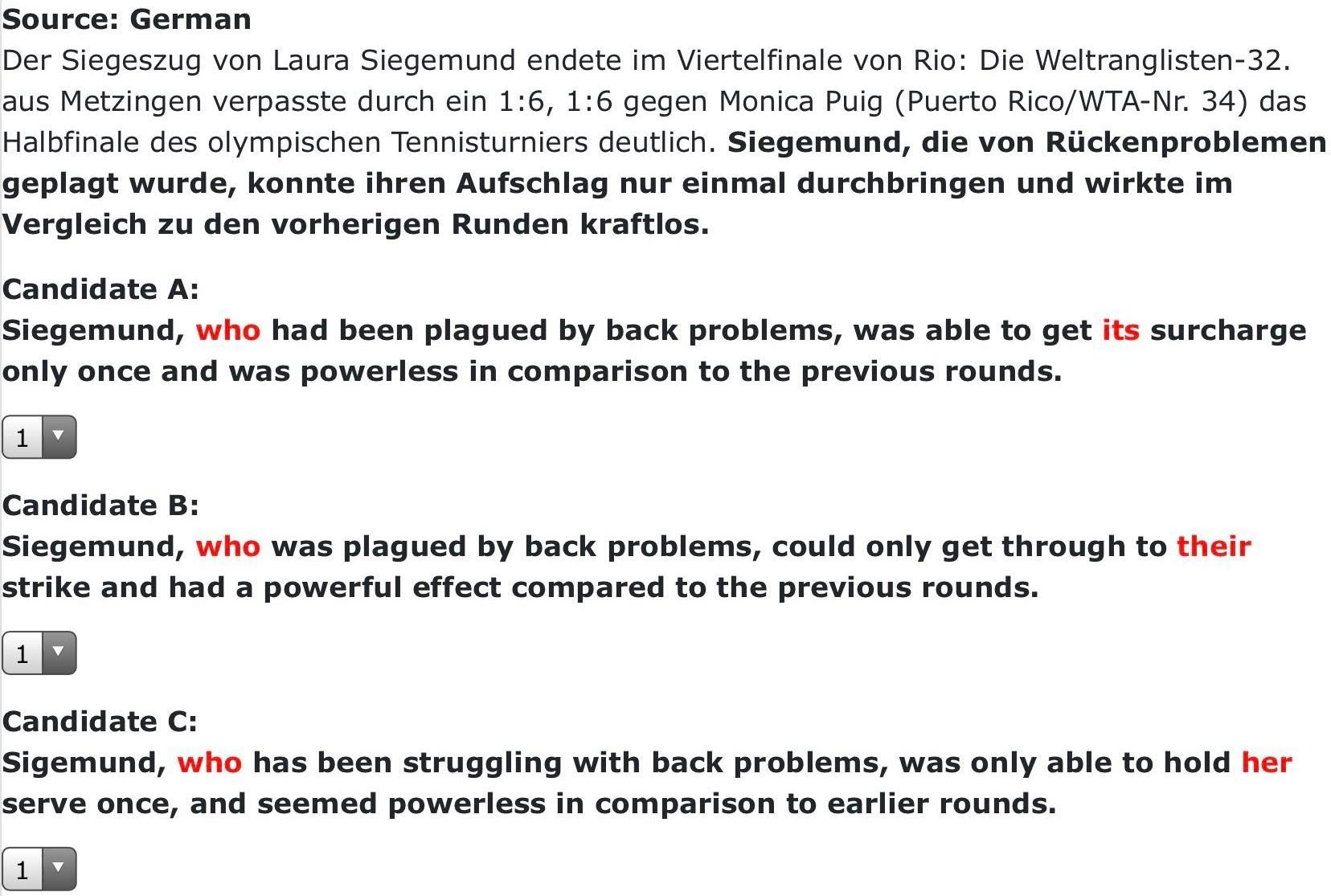}
    \vspace{-0.5em} %96
    \caption{User study interface (bilingual setup) for pronoun translation ranking. Pronouns in the sentences are highlighted in red.}
    \label{fig:mono_prn_rank}
\end{figure}

\paragraph{User Study.} The bilingual (German-English) user study interface for pronoun translation ranking is shown in Figure~\ref{fig:mono_prn_rank}.

\paragraph{Results.}
The total assigned scores (difference between reference score and translation score) obtained for each system after summing the over the samples in the respective testsets are given in Table \ref{table:anaphora-raw-scores}. The models are ranked based on these scores from lowest score (best performing) to highest score (worst performing).

\subsection{Coherence}
\label{app:coherence}

\paragraph{Re-trained model.}
We re-train the pairwise coherence model in \citet{unifiedcoherence} to suit the MT setting, with reference translations as the positive documents and the MT outputs as the negative documents. The results are shown in Table \ref{results:coherence-model}.

\begin{table}[h!]
\centering
\small
\tabcolsep=0.1cm
\begin{tabular}{l|c|c}  
\toprule
\textbf{Training data} & \textbf{Test data}  &  \textbf{Accuracy} \\
\midrule
WMT11-15 & WMT17-18 &  77.35\\
\bottomrule
\end{tabular}
\caption{Results of the re-trained coherence model.}
\label{results:coherence-model}
\end{table}

%\paragraph{User study.}
%Figure \ref{fig:coh_rank} shows our user study interface.

\paragraph{Results.}
The total assigned scores (difference between reference score and translation score) obtained for each system after summing the over the samples in the respective testsets are given in Table \ref{table:coherence-raw-scores}. The models are ranked based on these scores from lowest score (best performing) to highest score (worst performing).

\begin{table}[h!]
\centering
\small
\tabcolsep=0.1cm
{\begin{tabular}{l|c|c}  
\toprule
\multicolumn{3}{c}{De-En} \\
\toprule
\textbf{Rank} & \textbf{BLEU}  &  \textbf{Model Score} \\
\midrule
\sts & 31.65 &  109.0793 \\
\concat & \textbf{31.96}  & 206.7197 \\
\han & 29.22 & 211.4651 \\

\san & 29.32  & 214.5966 \\
\anap & 29.94  &  221.6826\\
\bottomrule
\end{tabular}}

{\begin{tabular}{l|c|c}  
\toprule
\multicolumn{3}{c}{Fr-En} \\
\toprule
\textbf{Rank} & \textbf{BLEU} &  \textbf{Model Score} \\
\midrule

\sts & 35.12 &  56.6195\\
\anap & 34.32 & 63.0893\\
\concat & \textbf{35.34}  &  118.7841 \\
\san & 33.48  &  128.7549\\
\han & 33.30 &  140.0488\\

\bottomrule

\end{tabular}}

{\begin{tabular}{l|c|c}  
\toprule
\multicolumn{3}{c}{Ru-En} \\
\toprule
\textbf{Rank} & \textbf{BLEU}  & \textbf{Model Score} \\
\midrule
\anap & \textbf{27.66}  & 132.7622 \\
\sts & 23.88 &  160.9542 \\
\han & 25.11 &  259.4806\\
\concat & 24.56  & 267.5505 \\
\san & 26.24  &  275.5277\\

\bottomrule
\end{tabular}}

\caption{Models ranked according to their performance (best to worst) in anaphora according to the evaluation model, with BLEU score for comparison. Model scores given here are obtained by subtracting the score for the model translation from the score for the reference translation, and summing the absolute score differences across the dataset. Hence, smaller model scores indicate better performance (closer to reference scores).} %\texttt{Syn} column is not counted within \texttt{Sum} column since \texttt{Syn} are not considered as errors for Anaphora. }
\label{table:anaphora-raw-scores}
\end{table}

\begin{table}[h!]
\centering
\small
\tabcolsep=0.1cm
\begin{tabular}{l|c|c}  
\toprule
\multicolumn{3}{c}{De-En} \\
\toprule
\textbf{Rank} & \textbf{BLEU} & \textbf{Coherence Score} \\
\midrule
\sts & 31.65 & 2179.468   \\
\san & 29.32 & 2185.442 \\
\concat & \textbf{31.96} & 2185.925 \\
\anap & 29.94 & 2280.091 \\
\han & 29.22  & 2393.837  \\

\bottomrule
\end{tabular}

\begin{tabular}{l|c|c}  
\toprule

\multicolumn{3}{c}{Fr-En}\\
\toprule
\textbf{Rank} & \textbf{BLEU} & \textbf{Coherence Score} \\
\midrule
\san & 33.48 & 248.355  \\
\sts & 35.12 & 250.472  \\
\concat & \textbf{35.34} & 254.779  \\
\anap & 34.32 & 272.604  \\
\han & 33.30  & 306.905  \\

\bottomrule
\end{tabular}

\begin{tabular}{l|c|c}  
\toprule
\multicolumn{3}{c}{Ru-En} \\
\toprule
\textbf{Rank} & \textbf{BLEU} & \textbf{Coherence score} \\
\midrule
\sts & 35.12 & 3617.631\\ %& 0.08   \\
\anap & 34.32 & 3753.255\\%& 0.14  \\
\concat & \textbf{35.34} & 4016.302\\ %& 0.16  \\
\san & 33.48 & 4259.422\\ % & 0.2  \\
\han & 33.30 & 4327.069\\% & 0.19  \\

\bottomrule
\end{tabular}

\caption{Models ranked according to their performance (best to worst) in coherence according our evaluation, with BLEU for comparison. Coherence scores given here are obtained by subtracting the score for the model translation from the score for the reference translation, and summing the absolute score differences across the dataset. Hence, smaller model scores indicate better performance (closer to reference scores). }

%with respect to total number of sentences in the texts within the coherence benchmark. (6 DE texts of 60 sentences total, 6 FR text of 38 sentences total)}
\label{table:coherence-raw-scores}
\end{table}

\subsection{Discourse Connectives}
\label{app:connectives}

% \begin{table}[h!]
% \centering
% \small
% \tabcolsep=0.1cm
% \begin{tabular}{l|c|c|c|c}  
% \toprule
% \textbf{Training data} & \textbf{Test data}  &  \textbf{P} &  \textbf{R} &  \textbf{F1} \\
% \midrule
% Sections 0-21 & Section 23 & 94.72 & 92.73 & 93.71 \\
% \bottomrule
% \end{tabular}
% \caption{Results of the explicit connective classifier.}
% \label{results:conn-classifier}
% \end{table}

\paragraph{Connective Classification model.}
We build an explicit connective classifier to identify candidates that are acting in the capacity of a discourse connective. The model consists of an LSTM layer \cite{hochreiter1997long} followed by a linear layer for binary classification, initialized by ELMo embeddings \cite{PetersELMo:2018}. {We use annotated data from the Penn Discourse Treebank (PDTBv3) \cite{PDTBv3} and conduct cross-validation experiments across all 25 sections. Our classifier achieves an average cross-validation precision of \textbf{95.58}, recall of \textbf{92.35} and F1 of \textbf{93.92}, which shows that it generalizes very well. The high precision also provides certainty that the model is classifying discourse connectives reliably. }

\begin{figure}
    \centering
    \includegraphics[scale=0.25]{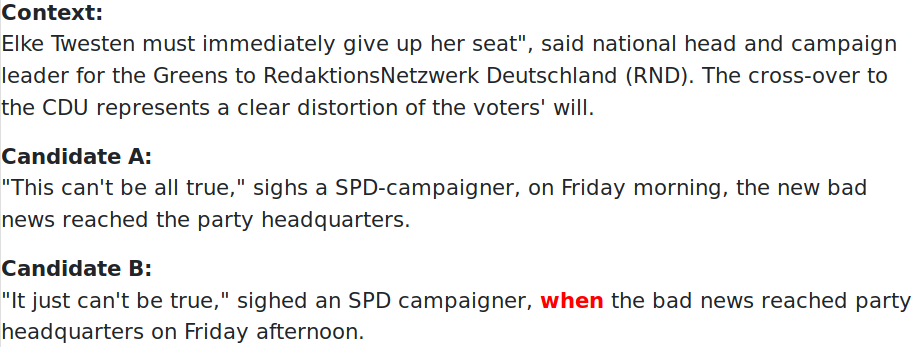}
    \caption{User study interface for reference vs. translation  missing connective study.}
    \label{fig:conn_noisy_study}
\end{figure}

\paragraph{User Study.}
For discourse connectives, we conducted two user studies. The first study in which the participants chose between the reference and its noisy version with the connective deleted was reported in the main paper. We present the connective-wise breakdown in Table \ref{results:conn-wise-noise}.

In the second study, the participants were shown the reference along with the system translation that was missing the connective (Figure \ref{fig:conn_noisy_study}). In this study, the setup has no artificially constructed data; the idea is to check if there is a possibility that the system translation is structured in such a way as to require no connective. However, the AC1 agreement for preferring the reference was \textbf{0.82} (2 annotators; different annotators from the first study) for this study as well, which is still quite high. Table \ref{results:conn-wise-sys} has the connective-wise breakdown; here we see that the results are slightly different for certain connectives, but overall the strong preference for the reference with the connective is retained. Our assumption that connectives must be translated is validated through both studies.

Note that for both studies, participants were also given options to choose `Neither' in case they didn't prefer either choice, or `Invalid' in case there was an issue with the data itself (\eg\, transliteration issues, etc.); data that was marked as such was excluded from further consideration.

\begin{table}[ht]
\centering
\small
\tabcolsep=0.1cm

\begin{tabular}{l|c|c|c|c}  
\toprule
\multicolumn{5}{c}{\textbf{Study 1: Reference vs. Connective Deleted Reference}}\\
\midrule
\textbf{Connective} & \textbf{AC1 Agr.}  &  \textbf{\# Ref} &  \textbf{\# Noisy } & \textbf{\# Tie}  \\
\midrule
 and & 0.96 & 136 & 4 &  11\\
 also & 1.0 & 35 & 3 & 18\\
 when & 1.0 &  29 & 0 & 0\\
 after & 1.0 & 23 & 0 & 0\\
 by & 1.0 & 12 & 0 & 2\\
 or & 1.0 & 6 & 0 & 0 \\
 as & 1.0 & 9 & 0 & 1 \\
 while & 1.0 & 9 & 0 & 3 \\
 so & 1.0 & 1 & 0 & 1\\
 because & 1.0 & 10 & 0 & 0\\
 then & 1.0 & 6 & 0 & 5 \\
 with & 1.0 & 5 & 0 & 1\\
 if & 1.0 & 4 & 0 & 0\\
 thus & 1.0 & 2 & 0 & 0 \\
 indeed & 1.0 &  0 & 0 & 2\\
 still & 1.0 &  2 & 0 & 2\\
 without & 1.0 & 2 & 0 & 0\\
 unless & 1.0 & 2 & 0 & 0\\
 until & 1.0 & 2 & 0 & 0 \\
 therefore & -0.33 &  1 & 0 & 1\\
 subsequently & 0 & 0 & 0 & 2\\
 ultimately & 0 &  0 & 0 & 2\\
 before & 1.0 & 8 & 0 & 0 \\
 previously & 0 & 0 & 0 & 2 \\
 once & 1.0 & 2 & 0 & 0 \\
 however & 1.0 & 2 & 0 & 0 \\
 in & 1.0 & 2 & 0 & 0 \\

\bottomrule
\end{tabular}
\caption{Connective-wise results for the user study with noisy data. The table also shows the number of times the \textbf{Ref}erence / \textbf{Noisy} translation was chosen (summed for both annotators). The \textbf{Tie} column shows the number of times the users showed no preference. Note that ties are not included in the agreement. Other samples not included were the ones marked as \textit{invalid} by the annotators due to misalignment errors, severe grammatical issues, etc.}
\label{results:conn-wise-noise}
\end{table}

\begin{table}[ht]
\centering
\small
\tabcolsep=0.1cm
\begin{tabular}{l|c|c|c|c}  
\toprule
\multicolumn{5}{c}{\textbf{Study 2: Reference vs. Missing Connective Translation}}  \\
\midrule
\textbf{Connective} & \textbf{AC1 Agr.}  &  \textbf{\# Ref} &  \textbf{\# Sys } & \textbf{\# Tie}  \\
\midrule
 and & 0.84 & 127 & 20 &  26\\
 also & 0.82 & 36 & 5 & 3\\
 when & 0.88 &  22 & 1 & 4\\
 after & 0.81 &  15 &  1 & 6\\
 by & 1.0 & 12 & 0 & 0\\
 or & -0.38 & 2 & 1 & 3 \\
 as & 0.79 & 12 & 1 & 1 \\
 while & 1.0 &  11 & 0 & 1 \\
 so & 1.0 & 8 & 0 & 0\\
 because & 1.0 & 7 & 0 & 1\\
 then & 0.57 & 6 & 2 & 4 \\
 with & 1.0 & 5 & 0 & 1\\
 if & 1.0 & 3 & 0 & 1\\
 thus & 1.0 & 2 & 0 & 0 \\
 indeed & 1.0 &  2 & 0 & 0\\
 still & 1.0 &  2 & 0 & 0\\
 
 without & 1.0 & 2 & 0 & 0\\
 unless & 1.0 & 2 & 0 & 0\\
 until & 1.0 & 2 & 0 & 0 \\
 therefore & 1.0 &  2 & 0 & 0\\
 subsequently & 1.0 & 2 & 0 & 0\\
 ultimately & 1.0 &  2 & 0 & 0\\
 before & -0.38 & 1 & 1 & 4 \\
 previously & 0 & 1 & 0 & 1 \\
 once & 0 & 1 & 0 & 1 \\
 however & 0 & 0 & 1 & 1 \\

\bottomrule
\end{tabular}

\caption{Connective-wise results for the user study with system translations. The table also shows the number of times the \textbf{Ref}erence / \textbf{Sys}tem translation was chosen (summed for both annotators). The \textbf{Tie} column shows the number of times the users showed no preference. Note that ties are not included in the agreement. Other samples not included were the ones marked as \textit{invalid} by the annotators due to misalignment errors, severe grammatical issues, etc.}
\label{results:conn-wise-sys}
\end{table}

\subsection{Model Parameters}
\label{app:model_parameters}
Parameters used to train \sts, \concat, \anap, and \san\ models are displayed in Table \ref{config-table-1}, and parameters for \han\ in Table \ref{config-table-2}.

\subsection{Datasets}
\label{app:training_dataset}

Our trainset is a combination of Europarl \cite{TIEDEMANN12.463}, IWSLT \cite{cettoloEtAl:EAMT2012} and News Commentary datasets, the development set is a combination of WMT-2016 and older WMT data (excluding 2014). We test on WMT-2014 data. We tokenize the data using the Moses software\footnote{\url{https://www.statmt.org/moses/}}, lowercase the text, and apply BPE encodings\footnote{\url{https://github.com/rsennrich/subword-nmt/}} from \citet{Sennrich2016NeuralMT}. We learn the BPE encodings with the command \texttt{learn-joint-bpe-and-vocab -s 40000}.

\subsection{Error Examples}
\label{app:error_examples}
Examples for the different types of errors encountered across the tasks are given in Table \ref{tab:error-examples2}.

\begin{table}
\centering
%\small
%\tabcolsep=0.1cm
 \begin{tabular}{|l|c|}  
\hline
 \textbf{Parameters} & \textbf{Values} \\
\hline
\multicolumn{2}{|c|}{\textbf{Step 1: sentence-level NMT}} \\
\cline{1-2}
 -encoder\_type & transformer \\
 -decoder\_type & transformer \\
 -enc\_layers & 6 \\
 -dec\_layers & 6 \\
 -label\_smoothing & 0.1 \\ 
 -rnn\_size & 512 \\
 -position\_encoding & - \\
 -dropout & 0.1 \\
 -batch\_size & 4096 \\
 -start\_decay\_at & 20 \\
 -epochs & 20 \\
 -max\_generator\_batches & 16 \\
 -batch\_type & tokens \\
 -normalization & tokens \\
 -accum\_count & 4 \\
 -optim & adam \\
 -adam\_beta2 & 0.998 \\
 -decay\_method & noam \\
 -warmup\_steps & 8000 \\
 -learning\_rate & 2  \\
 -max\_grad\_norm & 0 \\ 
 -param\_init & 0 \\
 -param\_init\_glorot & - \\
 -train\_part sentences & - \\ 
\cline{1-2}
 \multicolumn{2}{|c|}{\textbf{Step 2: HAN encoder}} \\
\cline{1-2}
 \textit{others - see Step 1} & \textit{others - see Step 1} \\
 -batch\_size & 1024 \\
 -start\_decay\_at & 2 \\
 -epochs & 10 \\
 -max\_generator\_batches & 32 \\
 -train\_part & all \\
 -context\_type & HAN\_enc \\
 -context\_size & 3 \\
\cline{1-2}
 \multicolumn{2}{|c|}{\textbf{Step 3: HAN joint}} \\
\cline{1-2}
 \textit{others - see Step 1} & \textit{others - see Step 1} \\
 -batch\_size & 1024 \\
 -start\_decay\_at & 2 \\
 -epochs & 10 \\
 -max\_generator\_batches & 32 \\
 -train\_part & all \\
 -context\_type & HAN\_join \\
 -context\_size & 3 \\
 -train\_from & [HAN\_enc\_model] \\
\hline
\end{tabular}
\caption{Configuration parameters for training \han\ model, taken from the authors' repository \url{https://github.com/idiap/HAN_NMT/}}
\label{config-table-2}
\end{table}

\begin{table*}
\centering
%\small
%\tabcolsep=0.1cm
 \begin{tabular}{|l|l|c|}  
\hline
\textbf{Model} & \textbf{Parameters} & \textbf{Values} \\
\hline
\san & \multicolumn{2}{c|}{ \textbf{Step1: sentence-level}} \\
\cline{2-3}
& batch\_size & 6250 \\
& update\_cycle & 4 \\
& train\_steps & 200000 \\
\cline{2-3}
& \multicolumn{2}{c|}{\textbf{Step 2: context-aware Transformer}} \\
\cline{2-3}
& num\_context\_layers & 1 \\

\hline
\anap & --optimizer & adam \\
& --adam-betas & '(0.9, 0.98)' \\
& --clip-norm & 0.0 \\
& --lr-scheduler & inverse\_sqrt \\
& --warmup-init-lr & 1e-07 \\
& --warmup-updates & 4000 \\
& --lr & 0.0007 \\
& --min-lr & 1e-09 \\
& --criterion & label\_smoothed\_cross\_entropy \\
& --label-smoothing & 0.1 \\
& --weight-decay &0.0 \\
& --max-tokens & 1024 \\
& --update-freq & 32  \\
& --share-all-embeddings & - \\
& --max-update & 100000 \\
\hline
\concat & --optimizer & adam \\
& --adam-betas & '(0.9, 0.98)' \\
& --clip-norm & 0.0 \\
& --lr-scheduler & inverse\_sqrt \\
& --warmup-init-lr & 1e-07 \\
& --warmup-updates & 4000 \\
& --lr & 0.0007 \\
& --min-lr & 1e-09 \\
& --criterion & label\_smoothed\_cross\_entropy \\
& --label-smoothing & 0.1 \\
& --weight-decay & 0.0 \\
& --max-tokens & 4096 \\
& --update-freq & 8 \\
& --share-all-embeddings & - \\
& --max-update & 100000  \\
\hline
\sts & \textit{as in \concat} & \textit{as in \concat} \\
\hline
\end{tabular}
\caption{Configuration parameters for training \san, \anap, \concat, \sts\  models. Parameters of \anap\ are taken from the original paper \cite{voita-etal-2018-context} and parameters of \san\ are taken from the authors' repository: \url{https://github.com/THUNLP-MT/Document-Transformer} and user manual for the THUMT library which provides the basic Transformer model: \url{https://github.com/THUNLP-MT/THUMT/blob/master/UserManual.pdf}. Parameters which are not listed were left as default.}
\label{config-table-1}. 
\end{table*}

\begin{table*}[t]
%    \small
    \centering
    \scalebox{0.8}
  { \begin{tabular}
    {|l|l|}
\toprule
\textbf{ Phenomenon} & \textbf{Example} \\
 \midrule
 \multicolumn{2}{|c|}{\textbf{Anaphora}}  \\
 \midrule
 Gender Copy 
 & S: \textit{Mir wurde \textbf{diese Wohnung}}  \textit{in Earls Court gezeigt, und \textbf{sie} hatte ...}  \\
 & T: I was shown this \textbf{apartment} in Earls Court , and \textbf{she} had ..  \\
 \midrule
 Named Entity 
 & T: \textit{... \textbf{Lady Liberty} is stepping forward.} \textit{\textbf{It} is meant to be carrying the torch of liberty}  \\
 & R: \textbf{She} is meant to be carrying the torch of Liberty. \\
 \midrule
 Language Specific 
%  & German pronoun \textit{ihr} can mean \textit{her}, \textit{their}, or \textit{your}; \textit{sie} can mean \textit{she} or \textit{you}  \\ 
 & S: \textit{Ihr Auftraggeber: Napoleon.}, the pronoun \textit{ihr} refers to the noun \textit{Karten} (English: \textit{maps}). \\
 & The German pronoun \textit{ihr} can mean \textit{her}, \textit{their}, or \textit{your}. \\
 & T: \textit{(..) detailed maps for towns and municipalities (...). \textbf{Your} contractor : Napoleon.} \\
 & R: (..) detailed maps for towns and municipalities (...). \textbf{Their} commissioner: Napoleon. \\
 \midrule
 \multicolumn{2}{|c|}{\textbf{Lexical Consistency}}  \\
\midrule
 Synonym 
%  & Words exchanged for synonyms (\textit{poll} - \textit{survey}, \textit{argument} - \textit{dispute}, \textit{association} - \textit{club})  \\
 & T: \textit{Watch the Tory party \textbf{conference}. The \textbf{convention} is supposed to be about foreign policy, (...).} \\
 & R: Under tight security - the Tory party \textbf{conference}. The party \textbf{conference} was to address foreign policy (...). \\
 \midrule
 Related Word 
%  & Words exchanged for hypernyms or hyponyms (\textit{ambulance} - \textit{car})  \\
%  & or related concepts (\textit{air traffic controller} - \textit{air navigation} - \textit{pilot}, \textit{wine} - \textit{vineyard})  \\
  & T: \textit{In the collision of the \textbf{car} with a taxi, a 27-year-old passer was fatally injured.} \\ 
  & R: A 27-year old passenger was fatally injured when the \textbf{ambulance} collided with a taxi. \\

  \midrule
  Named Entity 
%   &  \textit{Füchtorf} (name of a town) translated to \textit{feeding-community} \\
  & T: \textit{The \textbf{Feeding-Community} farmer , however , also had the ready-filled specialities.} \\
  & \textit{The demand for the good "made in \textbf{Feed orf}" was correspondingly high.} \\
  & R: But the \textbf{Füchtorf} farmer also had bottled specialties with him. \\
  & There was a lot of demand for the good "made in \textbf{Füchtorf}" beverage. \\
  \midrule
  Omission 
%   & Words omitted from the lexical chain \\
  & T: \textit{(...) during the single-family home attempt, it stayed by the royal highlands thanks to the burglar \textbf{alarm}.} \\
  & \textit{They got off when the culprits turned hand on Friday just before 20 a.m.} \\
  & R: It is thanks to the \textbf{alarm} system that the attempt in the Königswieser Straße at the single family home (...). \\ 
  & On Friday just before 20.00 the \textbf{alarm} rang when the offenders took action. \\ 
  \midrule
 \multicolumn{2}{|c|}{\textbf{Coherence}}  \\
  \midrule
 Ungrammatical 
 & T: \textit{"They didn't play badly for long periods -- like Stone Hages , like Hip Horst -- Senser.} \\ 
 & \textit{Only the initial phase, we've been totally wasted", \textbf{annoyed the ASV coach.}} \\
 & R: "Over long periods, they had - as in Steinhagen, as against Hüllhorst - not played badly. \\
 & We only overslept the initial phase", \textbf{said the ASV coach annoyed.} \\

 \midrule
 Hallucination 
 & T: \textit{Before appointing Greece , Jeffrey Pyett was the US ambassador to Kiev.} \\
 & \textit{When it came to the Maidan and the coup in 2014 , it was a newspaper. } \\
 & R: Before his appointment, Geoffrey Ross Pyatt was an ambassador in Kiyv.\\
 & During his mission, the Maydan events and state coup happened, reminds Gazeta.Ru \\
  \midrule

 Inconsistency  
 & T: \textit{The one-in-house airline crashed on Sunday afternoon at a parking lot near \textbf{Essen-Mosquitos}.} \\
 & \textit{\textbf{Essen Mill} is a small airport that's used a lot by airline pilots.} \\
 
 & R: On Sunday afternoon, the single-seated aircraft crashed (..) a parking lot near the airport  \textbf{Essen-Mülheim} \\
 & \textbf{Essen-Mülheim} is a small airport, which is frequently used by pilots with light private planes. \\

 \midrule
 \multicolumn{2}{|c|}{\textbf{Discourse Connectives}}  \\
\midrule
Omission 
& T: \textit{Two people died driving their car against a tree .}\\
& R: Two people died \textbf{after} driving their car into a tree.  \\
\midrule

Synonym 
& T: \textit{Naldo's "great sport" on the bank, \textbf{too}.} \\
& R:  \textit{Naldo is \textbf{also} a great athlete on the bench} \\

\midrule
Mistranslation 
& T: \textit{Gfk's leadership departs from disappointing business figures} \\
& R: GfK managing director steps down \textbf{after} disappointing figures \\

 \bottomrule
    \end{tabular}}

    \caption{Examples for the types of errors found in the translations. \textbf{S:} denotes source, \textbf{T:} denotes model translations while \textbf{R:} denotes  reference translations. }
    
    \label{tab:error-examples2}
\end{table*}

\end{document}